\documentclass[journal]{IEEEtran}

\usepackage{cite}
\usepackage{amssymb,amsfonts}
\usepackage{textcomp}
\usepackage[pdftex]{graphicx}
\graphicspath{{../pdf/}{../jpeg/}}
\DeclareGraphicsExtensions{.pdf,.jpeg,.png}

\usepackage{amssymb} 
\usepackage{algorithm}
\usepackage{algorithmicx}
\usepackage{algpseudocode}
\usepackage{dblfloatfix}

\usepackage[natural]{xcolor}
\usepackage[cmex10]{amsmath}
\usepackage[tight,footnotesize]{subfigure}
\usepackage{url}
\usepackage[nolist]{acronym}

\usepackage[textsize=scriptsize]{todonotes} 

\usepackage{booktabs}
\usepackage{multirow}
\usepackage[binary-units=true]{siunitx}

\DeclareSIUnit\pixel{px}
\DeclareSIUnit\fps{fps}
\usepackage{adjustbox}
\usepackage{rotating}
\usepackage{cleveref}
\Crefname{figure}{Fig.}{Figs.}
\crefname{figure}{Fig.}{Figs.}
\crefname{equation}{Eq.}{Eqs.}
\creflabelformat{equation}{#2\textup{#1}#3}
\crefname{table}{Tab.}{Tab.}
\Crefname{section}{Sec.}{Sec.}
\crefname{section}{Sec.}{Sec.}

\usepackage{leftidx}
\usepackage{multirow}

\usepackage{etoolbox}           
\renewcommand{\bfseries}{\fontseries{b}\selectfont} 
\robustify\bfseries             
\newrobustcmd{\B}{\bfseries}    

\usepackage{mathtools}
\DeclarePairedDelimiterX\set[1]\lbrace\rbrace{\def\given{\;\delimsize\vert\;}#1}
\DeclarePairedDelimiterX\setsplit[1]\lbrace\rbrace{\def\given{\;\delimsize\vert\;}#1}

\begin{document}
	
	\title{Pose and Semantic Map Based Probabilistic Forecast of Vulnerable Road Users' Trajectories}
	
	
	\author{Viktor Kress, Fabian Jeske, Stefan Zernetsch, Konrad Doll,~\IEEEmembership{Member,~IEEE,} and Bernhard Sick,~\IEEEmembership{Member,~IEEE}
		\thanks{V. Kress, F. Jeske, S. Zernetsch, and K. Doll are with the Faculty of Engineering,
			University of Applied Sciences Aschaffenburg, Aschaffenburg, Germany
			{\tt\footnotesize viktor.kress@th-ab.de, fabian.jeske@th-ab.de,
				stefan.zernetsch@th-ab.de, konrad.doll@th-ab.de}}
		\thanks{B. Sick is with the Intelligent Embedded Systems Lab, University of Kassel,
			Kassel, Germany
			{\tt\footnotesize bsick@uni-kassel.de}}
	}
	
	\maketitle
	
	\begin{abstract}
In this article, an approach for probabilistic trajectory forecasting of vulnerable road users (VRUs) is presented, which considers past movements and the surrounding scene. Past movements are represented by 3D poses reflecting the posture and movements of individual body parts. The surrounding scene is modeled in the form of semantic maps showing, e.g., the course of streets, sidewalks, and the occurrence of obstacles. The forecasts are generated in grids discretizing the space and in the form of arbitrary discrete probability distributions. The distributions are evaluated in terms of their reliability, sharpness, and positional accuracy. We compare our method with an approach that provides forecasts in the form of Gaussian distributions and discuss the respective advantages and disadvantages. Thereby, we investigate the impact of using poses and semantic maps. With a technique called spatial label smoothing, our approach achieves reliable forecasts. Overall, the poses have a positive impact on the forecasts. The semantic maps offer the opportunity to adapt the probability distributions to the individual situation, although at the considered forecasted time horizon of \SI[round-mode=places,round-precision=2]{2.52}{\second} they play a minor role compared to the past movements of the VRU. Our method is evaluated on a dataset recorded in inner-city traffic using a research vehicle. The dataset is made publicly available.

\end{abstract}
	
	%
	\IEEEpeerreviewmaketitle
	

	
	\section{Introduction}
\label{sec_introduction}
\subsection{Motivation} \label{subsec_motivation}
\begin{figure}
	\centering
	\begin{adjustbox}{clip,trim=0.0cm 0.0cm 0.0cm 0.0cm}\includegraphics[width=1\columnwidth]{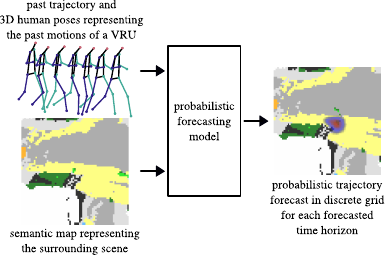}\end{adjustbox}
	\caption{Illustration of the approach for probabilistic trajectory forecasting. The model bases its forecasts on the VRU's past motions, represented by 3D poses, and semantic maps reflecting the surrounding scene. The colors correspond to the following semantic classes: \textit{static obstacle}~(black), \textit{dynamic obstacle}~(orange), \textit{sidewalk}~(yellow), \textit{road}~(dark grey), \textit{walkable vegetation}~(light blue), \textit{person}~(light green), \textit{unkown obstacles}~(dark green), and \textit{unkown free space}~(light grey). At the time of forecasting, the pedestrian is in the center of the map. In this scene, the pedestrian is walking on the sidewalk. For certain forecasted times, a discrete probability distribution (represented by the color gradient from red to blue on the right) is forecasted, expressing the likely future location of the VRU.}
	\label{fig_visual_abstract}
\end{figure}

In the future, automated systems will operate in areas shared with humans and they must understand human behavior to make interactions safe, efficient, and comfortable. This is particularly important for automated vehicles and vulnerable road users~(VRUs) in road traffic. A safe path planning of such vehicles requires a forecast of behavior and future trajectories of VRUs. However, future behavior is inherently fraught with uncertainty. Therefore, this work deals with probabilistic trajectory forecasting of VRUs, such as pedestrians and cyclists. The goal is the forecast of reliable probability distributions representing the movement capabilities of the VRUs fitted to the respective situation. The most important indicator for future behavior is the past movement. Besides the past trajectory, which is commonly used in the trajectory forecasting literature, we also include the body posture and movements of individual body parts. We address this by investigating the use of so-called 3D poses describing the three-dimensional positions of numerous joints along the body. Apart from past movements, the surrounding scenes of the VRUs is decisive for the future trajectory. It involves other road users, such as pedestrians, vehicles, static obstacles, lanes, and sidewalks. To encode such information, we use top-view semantic maps and consider them during forecasting. The overall approach is illustrated in~\cref{fig_visual_abstract}. All the data used for trajectory forecasting is obtained solely using a stereo camera and LiDAR mounted on a research vehicle and is generated fully automatically. Such sensors could be incorporated into ordinary vehicles in the near future. In literature, approaches for forecasting continuous as well as discrete probability distributions, can be found. We compare methods of both types to explore their strengths and weaknesses. Given the importance of the forecasted probability distributions, e.g., for path planning of automated vehicles, we focus on evaluating the distributions in terms of their reliability and sharpness.

\subsection{Related Work} \label{subsec_related_work}
This article mainly investigates probabilistic trajectory forecasting of VRUs while considering poses and surrounding scenes. By now, numerous works have been published regarding deterministic trajectory forecasting of VRUs in road traffic. For example, Keller and Gavrila~\cite{Keller.2014} forecasted trajectories of pedestrians potentially crossing the street by using Gaussian process dynamical models (GPDMs). Goldhammer et al.~\cite{Goldhammer.2019} used polynomial approximations of past velocities combined with a multi-layer perceptron (PolyMLP) to forecast the future trajectories of pedestrians and cyclists. However, forecasts are inevitably subject to uncertainty. An estimation of uncertainty is useful, for example, for safe path planning of automated vehicles. Without taking a negligible risk, an efficient traffic flow is not possible in areas shared by VRUs and automated vehicles. At the same time, path planning must consider the movement possibilities of VRUs and the risk taken must be quantifiable. Hence, methods for trajectory forecasting, including an estimation of the uncertainty, have been developed. For this purpose, some research forecasted several possible trajectories. Gupta et al.~\cite{Gupta.2018} used generative adversarial networks (GANs) and a pooling mechanism to incorporate dependencies among multiple people for forecasting multiple socially acceptable trajectories. By socially acceptable, the authors mean compliance with social norms, such as respecting a certain distance. A varying number of future trajectories were forecasted by memory augmented neural networks in \cite{Marchetti.2020}. However, such a set of forecasts only represents the uncertainty to a limited extent, and the quality of the uncertainty estimate is usually not evaluated. Instead, it is also feasible to forecast probability distributions that are either continuous or discretized regarding the spatial locations. Alahi et al.~\cite{Alahi.2016} focused on crowed spaces by forecasting bivariate Gaussian distributions for several pedestrians simultaneously via pooling based long short term memory (LSTM). The authors trained this model by minimizing the negative log-likelihood. Gaussian distributions describing the cyclists' future positions were forecasted in~\cite{pool.2019} using recurrent neural networks. In~\cite{Eilbrecht.2017}, forecasted trajectories were extended by an uncertainty estimation through an unconditional, constant model and used for path planning of autonomous vehicles. The authors of~\cite{Koschi.2018} proposed a method for forecasting all possible future positions of pedestrians in a set-based fashion using reachability analysis, contextual information, and traffic rules. The uncertainties of trajectory forecasts of cyclists were estimated in~\cite{Zernetsch.2019} in the form of unimodal Gaussian distributions with the help of a multi-layer perceptron (MLP). In addition, an approach to evaluate the reliability of the forecasted uncertainties was presented. Overall, this approach was able to make reliable forecasts for the motion types \textit{start}, \textit{stop}, \textit{turn left}, and \textit{turn right}, while the forecasts for the motion types \textit{move} and \textit{wait} were underconfident. Bieshaar et al.~\cite{Bieshaar.2020} extended single-output quantile regression to multivariate-targets for trajectory forecasting of cyclists. These so-called quantile surfaces represent star-shaped distributions using discrete quantile levels.

Besides forecasts of continuous distributions, distributions in discretized space are feasible as well. For example, Markov chains were used in~\cite{Wu.2018} to forecast probabilistic distributions as occupancy grids considering the pedestrians' potential goals. Jain et al.~\cite{Jain.2019} encoded the past in the form of multi-channel images and forecasts trajectories over long time horizons using a discrete residual flow network in the form of probability grids for each forecasted time step.

In addition to the past trajectory of the respective VRU, the posture and movements of individual body parts can be represented by 3D body poses and considered in trajectory forecasting. Quintero et al.~\cite{Quintero.2019} performed trajectory forecasting of pedestrians based on 3D poses with several balanced GPDMs. They were trained on 3D poses for different motion types, while the most similar model for the individual pedestrian behavior was chosen for trajectory forecasting. In~\cite{Kress.2019c}, 3D poses estimated from a moving vehicle were used for trajectory forecasting of pedestrians and cyclists. While both approaches achieved improvements by using poses, they did not include an estimate of the uncertainty.

The surrounding scenes, e.g., obstacles, lanes, sidewalks, or vegetation, influence the future trajectories of VRUs. Accordingly, numerous methods considered the surrounding for trajectory forecasting: Xue et al.~\cite{Xue.2018} used an LSTM architecture for trajectory forecasting of pedestrians with an upstream convolutional neural network~(CNN) extracting relevant features from top-view images of the scene. In~\cite{Cheng.2020}, the surrounding was expressed through motion heat maps representing the prior movements of other VRUs, segmented maps describing the accessible areas, and aerial photography images. This representation is processed by a CNN and, together with the past trajectory and data about nearby road users, used to forecast several possible trajectories by means of a neural network. Ridel et al.~\cite{Ridel.2020} performed a semantic segmentation on top-view images using residual networks (ResNet). They encoded the past trajectory in binary 2D grids and forecasted discrete probability distributions in agent-centric grids for future time steps using a network architecture consisting of CNNs and convolutional LSTMs~(ConvLSTM). However, these approaches require current images from infrastructure-based cameras or drones, which is usually not feasible in road traffic.
In~\cite{Jain.2019}, images encoding semantic maps were used for trajectory forecasting of pedestrians. The images mask information of maps, e.g., crosswalks, drivable surfaces, lanes, and traffic light states, which were annotated semi-automatically, and the history of dynamic objects detected from LiDAR and camera. The maps were centered on the respective pedestrian and processed by a CNN. Marchetti et al.~\cite{Marchetti.2020} projected semantic labels of static objects aggregated over time into a top-view map using a LiDAR point cloud and IMU data from a moving vehicle. Those maps were subsequently used to refine the forecasts to ensure compatibility with the surrounding.

\subsection{Main Contributions and Outline of this Article} \label{subsec_main_contrib}
Our main contributions are the following: First, we combine the past trajectory with 3D poses of pedestrians and cyclists and maps of the surrounding scenes for discrete probabilistic trajectory forecasting by means of neural networks. The 3D poses represent the past motions of different body parts in great detail. To the best of our knowledge, this is the first time poses are used for probabilistic trajectory forecasting. We use semantic maps reflecting both the static (e.g., sidewalks or obstacles) and dynamic surrounding (e.g., vehicles or other VRUs) to align the forecasts with the surroundings. We individually examine the impact of the past trajectory, 3D poses, and maps on trajectory forecasting.

Second, we analyze the forecasted distributions in detail with respect to their reliability, sharpness, and positional accuracy. We propose a method to calibrate the discrete probabilistic trajectory forecasting models regarding reliability and investigate the effects of the resolution of the discrete probability distributions on the quality of the forecasts, which has not been studied before in the literature.

Third, we evaluate the advantages and disadvantages of forecasting discrete versus continuous probability distributions. For this purpose, we use the approach from~\cite{Zernetsch.2019} as a continuous comparison method. Since this approach is originally based solely on the past trajectory, we extend it to include 3D poses as well.

Fourth, all used data were generated fully automatically with the help of a single vehicle's sensors in real road traffic. In contrast to the literature, neither drones, infrastructure-based sensors, nor manual annotations were used. Accordingly, the approach is appropriate for an application in automated vehicles. The dataset is made publicly available~\cite{Kress_Dataset.2021}.

The remainder of this article is organized as follows: \cref{sec_dataset} introduces the dataset, while in~\cref{sec_method} the discrete~(\cref{subsec_discrete_trajectory_forecast}) and continuous trajectory forecasting methods~(\cref{subsec_continuous_trajectory_forecast}) are explained. This sections focuses in particular on the network architecture, the training methodology, and evaluation metrics. Next, the results are presented in~\cref{sec_experimental_results}. It covers the process of hyperparameter optimization, reliability calibration using spatial label smoothing, the impact of poses, semantic maps and the cell size, and a comparison of the discrete and continuous forecasting approaches. Finally, we conclude with a summary and an outlook on future work~(\cref{sec_conclusion}).
	\section{Dataset} \label{sec_dataset}
The dataset was recorded with a moving vehicle in inner-city traffic in Aschaffenburg, Germany. The recordings were taken at different times of the day and year over four years and under different weather conditions to reflect road traffic variability. The driven routes cover different types of roads, such as multi-lane roads, roads with bike lanes, traffic-lighted and non-traffic-lighted intersections, traffic-calmed areas, crosswalks, bus stops, and more. Some of the recorded pedestrians and cyclists were instructed to walk or ride specific routes. The remaining VRUs were uninvolved and uninstructed individuals of all ages and agilities. The vehicle was equipped with a stereo camera behind the windshield, a LiDAR in the vehicle's front, and a vehicle localization system. The trajectories of all VRUs up to a distance of $\SI{25}{\m}$ were recorded using the stereo camera and a Kalman tracker. To obtain the 3D poses of the VRUs, first, the 2D poses, i.e., the two-dimensional coordinates of several joints in the images, were estimated using the CNN proposed in~\cite{Cao.2017} followed by a reconstruction of plausible 3D poses using the approach from~\cite{Tome.2017}. More details on this procedure and an evaluation of its accuracy can be found in~\cite{Kress.2018}. Compared to 2D poses or other image-based methods, 3D poses have the advantage of being independent of the perspective of the recording camera and allow for a compensation of the vehicle's own motion.

Further, semantic maps were generated describing the particular scene. For that purpose, semantic segmentation of the stereo camera images was performed using the approach from~\cite{Wu.2019}. To create a top-view semantic map, the semantic information about the surrounding was fused with an occupancy map based on the point cloud of the LiDAR. In this way, the high positional accuracy of the LiDAR is combined with the density of the camera images. The maps contain $n_s=8$ semantic classes: \textit{Static obstacles} include all types of non-movable barriers that are not negotiable by pedestrians and cyclists. Among these are, e.g., buildings, walls, fences, traffic signs, or vegetation. \textit{Dynamic obstacles}, on the other hand, comprise all types of moving or parked vehicles. The other classes are \textit{sidewalk}, \textit{road}, \textit{walkable vegetation}, such as meadows, and \textit{person} including pedestrians and cyclists. The remaining two classes are \textit{unknown obstacles} and \textit{unknown free space}. They cover areas for which the semantic class could not be determined. By aggregating all data in the dataset, a map of the driven roadways was created, containing only the invariant semantic classes. This initial map is subsequently enhanced at each point in time by the measurements of the immediate past resulting in an up-to-date map including moving elements. In a real-world application, the initial map could be replaced by commercially available high precision maps, and the current maps could be shared between multiple vehicles. The maps have a spatial resolution of $0.35\times\SI[round-mode=places,round-precision=2]{0.35}{\metre\squared}$ and a temporal frequency of $\SI{25}{\fps}$. Each pixel is classified as one of the eight classes.

The dataset contains $2351$~trajectories of pedestrians and $1075$~trajectories of cyclists with corresponding 3D poses and semantic maps. Each point in time of the trajectories was manually annotated with a motion type. The motion types comprise the states \textit{wait}, \textit{start}, \textit{move}, and \textit{stop} for pedestrians and cyclists. Additionally, a distinction is made between the motion types \textit{turn left} and \textit{turn right} for cyclists. These two motion types do not exist for pedestrians because they are difficult to annotate even for humans due to the agility of pedestrians. These annotated motion types are used solely for a differentiated evaluation in~\cref{subsec_results}. About $\SI{60}{\percent}$ each of the pedestrian and cyclist data is used as the training set, $\SI{20}{\percent}$ as validation set, and the remaining $\SI{20}{\percent}$ as the test set. It was ensured that the same location does not occur in different sets and that the motion types are distributed as equally as possible. This is important to guarantee the generalization ability to arbitrary locations and for a fair comparison with methods without knowledge of the surrounding. To ensure rotational invariance, the trajectories, poses, and semantic maps are randomly rotated and augmented by a factor of 3 through repeated random rotation.

The dataset including trajectories, 3D poses, semantic maps, and motion types has been made publicly available \cite{Kress_Dataset.2021}. All the data were recorded in accordance with the guidelines of the University of Applied Sciences Aschaffenburg and German privacy laws.

\section{Method}
\label{sec_method}
\subsection{Discrete Probabilistic Trajectory Forecast} \label{subsec_discrete_trajectory_forecast}
This work aims at forecasting probability distributions describing the possible future locations of the VRUs discretized in time and space. Therefore, we forecast the distribution~$\hat{p}_{t_f}$ for the forecasted time horizons $t_f\in T_f$ for which $\sum_{\vec{g}\in G}\hat{p}_{t_f}(\vec{g})=1$. Here, $G\subset\mathbb{R}^2$ is a grid centered on the current position of the respective VRU with cell size~$e^2$ discretizing the space and $\hat{p}_{t_f}(\vec{g})$ denotes the forecasted probability for a cell with center point $\vec{g}$.
\subsubsection{Input Features}
We train and analyze three models based on different feature sets for trajectory forecasting: The model based on the past trajectory, referenced hereafter by the abbreviation~d\_t~(\textbf{d}iscrete forecasting model based on past \textbf{t}rajectory), uses the VRU's two-dimensional head position of the ground plane for each time~$t_i$ in the observation period~$T_i$. Accordingly, the feature set is given by $f_{d\_t}=\set*{[x_{Head,t_i}, y_{Head,t_i}] \given t_i \in T_i}$. The origin of the corresponding coordinate system is the head position at the current time~$t_c$. 

The second model~d\_tp~(\textbf{d}iscrete forecasting model based on past \textbf{t}rajectory and \textbf{p}oses) additionally uses 3D poses as input. Instead of the head position, the feature vector for time~$t_i$ consists of the three-dimensional position of 13 joints along the body. A sequence of 3D poses with associated joint positions is shown in~\cref{fig_visual_abstract}. The coordinate system remains the same. All 3D poses are scaled in 3D space to have the same width of the hips. Again, the final feature set is obtained by concatenating the observation period: $f_{d\_tp}=\{[x_{Head,t_i}, y_{Head,t_i}, z_{Head,t_i}, ..., z_{LFoot,t_i}]\mid t_i \in T_i\}$. The feature sets of both models are normalized over all training samples using the statistical z-transformation.

Finally, the third model~d\_tpm additionally uses the semantic \textbf{m}ap~$m_{t_c}$ at current time~$t_c$ to adapt the trajectory forecasts to the particular scene. The origin of the semantic map is the current head position of the respective VRU. The semantic maps have the same dimensions and cell size as the forecasted grid~$G$. They are converted into multichannel images. Each channel represents a binary image encoding the presence or absence of one of the eight semantic classes mentioned above at each particular location.

\subsubsection{Network Architecture}
\begin{figure*}
	\centerline{\includegraphics[width=0.99\textwidth]{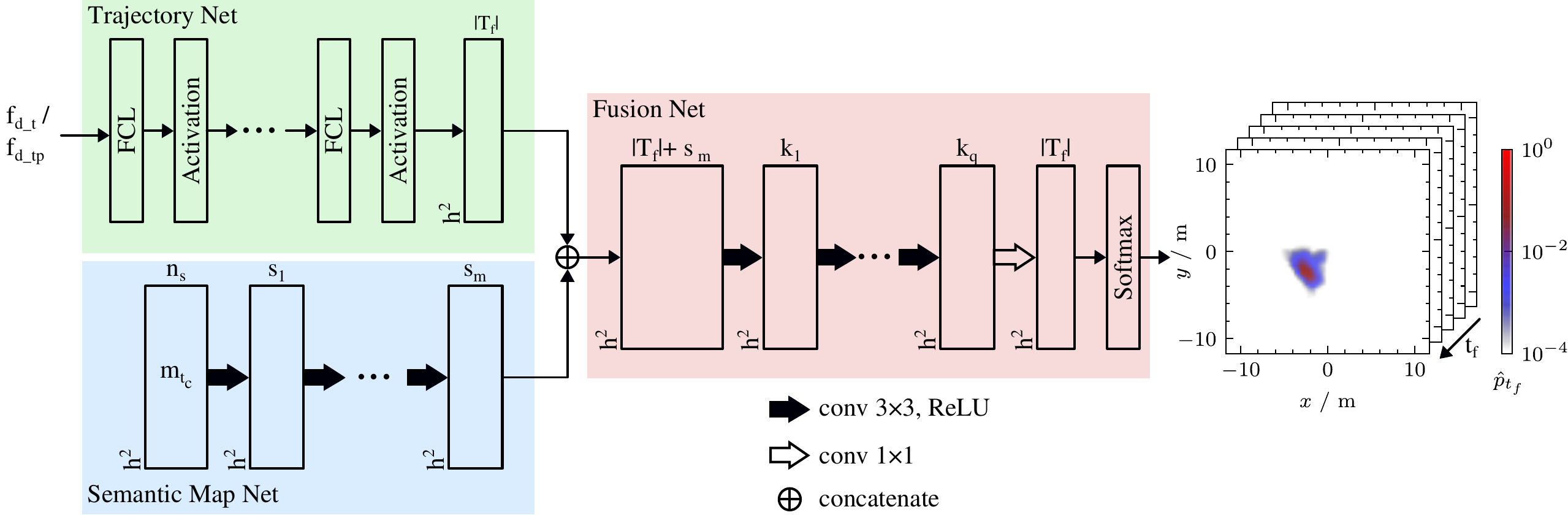}}
	\caption{Two stream network architecture with the so-called trajectory net processing the input trajectories and poses and the semantic map net receiving the semantic maps. Boxes represent the feature maps with their dimensions given in the lower-left corner and the number of channels on top of each box.}
	\label{fig_architecture}
\end{figure*}
The neural network has a two-stream architecture (\cref{fig_architecture}). The trajectory net is responsible for processing the past trajectory and the 3D poses. Accordingly, depending on the model, the feature sets $f_{d\_t}$ or $f_{d\_tp}$ serve as input for this stream. It consists of several fully connected layers, each followed by a Rectified Linear Unit (ReLU) activation function. The number of layers and the number of neurons per layer are hyperparameters chosen in an optimization step in \cref{subsec_hyper_parameter_optimization}. As a result, this network segment yields a feature map of size~$h^2$ with $|T_f|$ channels. The dimensions correspond to those of the final network output~$\hat{p}$. The second stream, called semantic map net, processing the semantic map consists of convolutional layers, each with a $3\times3$ filter and ReLU activation. This stream is omitted for the models d\_t and d\_tp, as they do not consider the semantic map. The two streams' feature maps are concatenated along the channel dimensions and processed in the so-called fusion net by further convolutional layers, again with $3\times3$~filter and ReLU activation. The number of convolutional layers and the number of filters of each convolutional layer are additional hyperparameters. Other activation functions, filter sizes, and the addition of pooling layers did not improve the forecasts. The last layer is a linear convolution with $1\times1$ filter producing a grid for each forecasted time horizon. Finally, a softmax activation function is applied for each grid to obtain the final probability distributions. We have also tested other network architectures, such as UNets~\cite{Ronneberger.2015} or ConvLSTMs~\cite{Shi.2015}, achieving similar results.
\subsubsection{Training}
\label{sec_training}
We train the models d\_t and d\_tp by minimizing the cross entropy~(\cref{eq:loss_tp}) between the forecasted distribution~$\hat{p}_{t_{f},i}$ and target distribution~$p_{t_{f},i}$ for all~$N$~training samples and forecasted time horizons~$T_f$.
\begin{equation}
L = -\frac{1}{|T_f|}\sum_{t_f\in T_f}\frac{1}{N}\sum_{i=1}^{N}\sum_{\vec{g}\in G} p_{t_f,i}(\vec{g})\log(\hat{p}_{t_f,i}(\vec{g}))
\label{eq:loss_tp}
\end{equation}
The cross entropy with one-hot encoded target tends to produce overconfident, i.e., to narrow, probability distributions that are not reliable. Typically, in order to avoid overconfidence, label smoothing with uniform distribution over the classes is applied. However, in our case, the grid cells are spatially related. Therefore, we use spatial label smoothing: Instead of a uniform distribution we use a Gaussian distribution~$p_{t_{f},i}(\vec{g})=\mathcal{N}_{\vec{y}_{t_{f},i},\sigma_{t_{f}}}(\vec{g})$ with the actual grid position~$\vec{y}_{t_{f},i}$ as expected value and standard deviation~$\sigma_{t_f}$ as target distribution. The standard deviations are additional hyperparameters. Their effects are examined in~\cref{subsec_calibration}.

For model d\_tpm the loss function~(\cref{eq:loss_tp}) is used to consider obstacles in the surroundings. For this purpose we define obstacles $c_{t_{f},i}(\vec{g})$ for sample~$i$ equaling 1 if there is an obstacle at the cell with center point~$\vec{g}$ and 0 otherwise. In this context, we declare obstacles as objects of the semantic classes \textit{static obstacles} and \textit{dynamic obstacles}. Thus, $c_{t_{f},i}$ can be derived from the semantic map~$m_{t_{f},i}$ at the respective time. Instead of the Gaussian distribution a modified Gaussian distribution $p_{t_{f},i}(\vec{g})=\mathcal{N}^{c_{t_{f},i}}_{\vec{y}_{t_{f},i},\sigma_{t_{f}}}(\vec{g})$ is used as target equaling 0 at locations with obstacles. The probabilities are scaled such that $\sum_{\vec{g}\in G}p_{t_f,i}(\vec{g})=1$.

\subsubsection{Evaluation Method}
\label{sec_eval_method}
Several scores describing different properties are used to evaluate the estimated probability distributions. They are introduced and explained in this section.
\paragraph{Reliability}
The reliability measures whether the variances of the distributions are estimated correctly. In \cite{Zernetsch.2019}, reliability is calculated by comparing confidence intervals of the forecasts with the observed frequency of actual positions within the interval. In the following, we use a similar approach adapted to the discrete probability distributions. In favor of clarity, we omit the index for the respective forecasted time horizon~$t_f$ below. The procedure is the same for all forecasted time horizons. We define a set of cell center points~$\Phi$ for which the associated cells have forecasted probabilities~$\hat{p}(\vec{g})$ greater than the forecasted probability at the cell of the actual position~$\hat{p}(\vec{y})$ for each sample~(\cref{eq:confidence_level_positions}). The respective confidence level~$C$ given the actual grid position~$\vec{y}$ and the forecast~$\hat{p}$ is obtained by adding the forecasted probabilities of all cells with the center points within the set~(\cref{eq:confidence_level}).
\begin{equation}
\Phi(\vec{y},\hat{p})=\set*{\vec{g}\in G \given \hat{p}(\vec{g})\geq \hat{p}(\vec{y})}
\label{eq:confidence_level_positions}
\end{equation}
\begin{equation}
C(\vec{y}, \hat{p}) = \sum_{\vec{g}\in \Phi(\vec{y},\hat{p})} \hat{p}(\vec{g})
\label{eq:confidence_level}
\end{equation}
The observed frequency~$f_o$ in the dataset with $M$ samples is calculated according to~\cref{eq:frequency_of_occurence} for a given confidence level~$1-\alpha$. A measure of reliability is obtained by comparing the observed frequencies to the given confidence levels. Ideally, they should be identical. A visualization is achieved by plotting the observed frequency given the confidence level, whereby the diagonal describes perfect reliability (see~\cref{fig_cal_rel_plots} for an example).
\begin{equation}
f_o(1-\alpha)=\frac{1}{M}\sum^{M}_{i=1}
\begin{cases}
1, & \text{if  } C(\vec{y}_i,\hat{p}_i) \leq 1-\alpha\\
0, & \text{otherwise}
\end{cases}
\label{eq:frequency_of_occurence}
\end{equation}
For a numerical evaluation of the reliability, we use a score similar to the widely used Expected Calibration Error~(ECE)~\cite{Naeini.2015} (\cref{eq:reliability_score}). Here, the range of possible confidence levels is partitioned into $B$~bins. The score is given by the averaged sum of the absolute difference between the observed frequency and given confidence level weighted by the number of forecasts~$j_b$ within the respective confidence level bin. It is averaged over all forecasted times~$T_f$, to gain a score for the entire forecasts.
\begin{equation}
\text{ECE}=\frac{1}{|T_f|M}\sum_{t_f\in T_f}\sum^{B}_{b=1}j_{b,t_{f}}|(1-\alpha_b)-f_{o,t_{f}}(1-\alpha_b)|
\label{eq:reliability_score}
\end{equation}
This score is related to the definition of reliability obtained by a decomposition of the Brier score~\cite{Murphy.1973} and is used for model calibration in~\cref{subsec_calibration}.

\paragraph{Sharpness}
In order to evaluate the sharpness of the forecasted distributions, we determine the area covered by a specific confidence level. For this purpose we search for a threshold~$\tau=\max(\xi)$ such that $\sum\set*{\hat{p}(\vec{g}) \given \hat{p}(\vec{g})\geq \xi}\geq 1-\alpha$. The area of a given confidence level and sample is then given by:
\begin{equation}
A(1-\alpha) = |\set*{\vec{g} \given \hat{p}(\vec{g})\geq \tau}|e^2
\label{eq:confidence_area}
\end{equation}
The sharpness~$S$~(\cref{eq:sharpness_score}) is defined as the mean area~$\overline{A}_{t_{f}}$ of all samples of the forecasted time horizon~$t_f$ normalized to the time horizon and averaged over all forecasted time horizons~$T_f$.
\begin{equation}
S(1-\alpha) = \frac{1}{|T_f|}\sum_{t_{f} \in T_{f}}\frac{\overline{A}_{t_{f}}(1-\alpha)}{t_f}
\label{eq:sharpness_score}
\end{equation}

\paragraph{Positional Accuracy}
The weighted average Euclidean error~(WAEE) is used to evaluate the positional accuracy~(\cref{eq:wAEE}). For a given forecasted time horizon, this metric determines the average sum of the $M$~Euclidean distances between each cell center point~$\vec{g}$ in the grid and the actual grid position~$\vec{y}_{i, t_{f}}$ weighted according to the forecasted probability~$\hat{p}_{i, t_{f}}$ for the cell with that center position.
\begin{equation}
\text{WAEE}_{t_{f}} = \frac{1}{M}\sum^{M}_{i=1}\sum_{\vec{g}\in G}\hat{p}_{i, t_{f}}(\vec{g}) {\lVert \vec{g}-\vec{y}_{i, t_{f}} \rVert}_2
\label{eq:wAEE}
\end{equation}
To gain a metric evaluating the forecasts over all forecasted time horizons, we calculate the average specific~WAEE~(ASWAEE). The metric normalizes and averages the WAEE according to the forecasted time horizon~(\cref{eq:ASWAEE}). It is the probabilistic equivalent of the average specific average Euclidean error introduced in~\cite{Goldhammer.2019}.
\begin{equation}
\text{ASWAEE} = \frac{1}{|T_f|}\sum_{t_f\in T_f}\frac{\text{WAEE}_{t_{f}}}{t_f}
\label{eq:ASWAEE}
\end{equation}

\paragraph{Forecasts of Obstacles Collisions}
The impact of using the semantic maps for trajectory forecasting is difficult to quantify, as discussed in~\cref{sec_impact_semantic_maps}. However, the proportion of the probability distribution located at areas occupied by static obstacles can be measured. This is obviously an incorrect forecast as long as the semantic maps are assumed to be correct. We define the so-called occupancy score $O_{t_f}$~(\cref{eq:occupancy_score}) for a certain forecasted time horizon as the sum of forecasted probabilities at locations of static obstacles~$o_{t_f,i}$. Here, $o_{t_f,i}$ is a binary map with value 1 at locations of static obstacles.
\begin{equation}
O_{t_f} = \frac{1}{M}\sum_{i=1}^{M} \sum_{\vec{g} \in G} \hat{p}_{t_f,i}(\vec{g})o_{t_f,i}(\vec{g})
\label{eq:occupancy_score}
\end{equation}
The metric refers only to static obstacles and not to dynamic ones, since their movements are again a matter of probability and therefore collisions of the forecasts with dynamic obstacles are not necessarily wrong.

\subsection{Continuous Probabilistic Trajectory Forecast} \label{subsec_continuous_trajectory_forecast}
We compare our approach for probabilistic forecasting of trajectories in discrete space with the method from~\cite{Zernetsch.2019} for forecasting continuous probability distributions. This method estimates the forecasts' uncertainty in the form of Gaussian distributions whose parameters (expected values, variances, correlation coefficients) are forecasted by a neural network with feed-forward architecture and fully connected layers. As input, the approach originally uses the past head positions of cyclists in an ego coordinate system. The ego coordinate system has its origin in the current position of the respective VRU, and the $x$- and $y$-axes are defined depending on the movement direction. We train and optimize the neural network on our dataset separately for pedestrians and cyclists. This model is referred to as~c\_t~(\textbf{c}ontinuous forecasting model based on past \textbf{t}rajectory). In addition, another model is created for pedestrians and cyclists, respectively, by extending the input feature space with poses (model~c\_tp). The poses are rotated according to the movement direction to obtain compatibility with the ego coordinate system. The feature set is defined by $f_{c\_{tp}}=\{\leftidx{^{ego}}{[x_{Head,t_i}, y_{Head,t_i}, z_{Head,t_i}, ..., z_{LFoot,t_i}]}\mid t_i \in T_i\}$. We refer to these comparison methods hereafter as continuous models. The evaluation of the continuous forecasting models is done equivalent to the discrete models using reliability, sharpness, and ASWAEE. However, the calculations are realized by sampling from the continuous probability distributions.

	\section{Experimental Results}
\label{sec_experimental_results}

\subsection{Hyperparameter Optimization}
\label{subsec_hyper_parameter_optimization}
\sisetup{round-mode=places,round-precision=2}
We train the models d\_t, d\_tp, d\_tpm, c\_t, and c\_tp separately for pedestrians and cyclists, resulting in a total of 10 models. For training the networks, the adaptive moment estimation~(Adam) optimizer~\cite{Kingma.2014} is used as well as early stopping as regularization technique.

We forecast probability distributions for five time horizons~$T_f=\{\SI{0.44}{\second}, \SI{0.96}{\second}, \SI{1.48}{\second}, \SI{2.0}{\second}, \SI{2.52}{\second}\}$ using an observation period of $\SI{1}{\second}$ $T_i=\{\SI{-0.96}{\second}, ..., \SI{-0.04}{\second}, \SI{0.00}{\second}\}$ corresponding to 25 position measurements. The size of the forecasted grid is defined such that the positions of all pedestrians or cyclists within our dataset are located certainly within the grid for all time horizons. For pedestrians the size is $h^2=67\times\SI{67}{\pixel\squared}$ and for cyclists $147\times\SI{147}{\pixel\squared}$ corresponding to $23.45\times\SI[round-mode=places,round-precision=2]{23.45}{\metre\squared}$ and $51.45\times\SI[round-mode=places,round-precision=2]{51.45}{\metre\squared}$, respectively. The size of the grids could be chosen smaller for short time horizons. However, this was not done in favor of a simpler network architecture and implementation. The size of a cell equals $e^2=0.35\times\SI[round-mode=places,round-precision=2]{0.35}{\metre\squared}$, which corresponds to the resolution of the semantic maps and approximately to the area occupied by a human being. However, since the cell size is principally a hyperparameter, its effects are examined in~\cref{subsec_cell_size}.

The hyperparameters are optimized by parameter sweeps using the validation dataset. Afterward, the networks obtaining the lowest ECE on the validation data are trained on a combined set of training and validation data and evaluated on the test dataset.
All hyperparameters of the network architecture are optimized only for pedestrians and adopted for cyclists to keep the computational cost reasonable. The resulting parameters of the discrete models are summarized in~\cref{tab_hyperparameters}. The final networks of models~c\_t and c\_tp each consist of 2 hidden layers with 100 neurons per layer.
\setlength{\tabcolsep}{7pt}
\begin{table}
	\caption{The selected hyperparameters of the network architecture. Note that, in addition to the specifications in the table, the fusion net always concludes with a convolutional layer with $1\times 1$ filter.}
	\begin{center}
		\begin{tabular}{@{}lcccr@{}}\toprule
			\bfseries parameter & \bfseries d\_t & \bfseries d\_tp & \bfseries d\_tpm \\\midrule
			number of hidden layers in trajectory net & 4 & 5 & 5 \\
			number of neurons in trajectory net & 150 & 50 & 50 \\
			number of conv in semantic map net & - & - & 1 \\
			number of filters in semantic map net & - & - & 8 \\
			number of conv in fusion net & 2 & 2 & 2 \\
			number of filters in fusion net & 10 & 20 & 20 \\\bottomrule
		\end{tabular}
		\label{tab_hyperparameters}
	\end{center}
\end{table}

\subsection{Reliability Calibration}
\label{subsec_calibration}
\sisetup{round-mode=places,round-precision=3}
One focus of this work is the forecast of reliable probability distributions, as they have a safety-relevant meaning, e.g., for path planning of autonomous vehicles. However, the use of the cross entropy as loss function leads to overconfident distributions, which are critical for an application, e.g., path planning of automated vehicles. The left plot in~\cref{fig_cal_rel_plots} shows the reliability diagram for the model~d\_tp for pedestrians trained using the cross entropy evaluated on the validation dataset. The curves fall below the ideal diagonal for all forecasted time horizons, which characterizes an overconfident distribution. There are considerable differences in the reliability of the different forecasted time horizons: For increasing time horizons, the reliability approaches more and more the diagonal. Overall, an ECE of $\SI[round-mode=places,round-precision=1]{8.590476610776}{\percent}$ is achieved. A popular approach to calibration is temperature scaling~\cite{Guo.2017}. Therefore, we performed temperature scaling separately for each forecasted time horizon on the validation dataset. While the reliability indeed improves with an ECE of~$\SI[round-mode=places,round-precision=1]{6.3898381056433
}{\percent}$, spatial label smoothing as described in~\cref{sec_training} achieves better results on our dataset. Here, the ECE is reduced to $\SI[round-mode=places,round-precision=1]{4.6495028272155}{\percent}$. The corresponding reliability diagram is shown on the right in~\cref{fig_cal_rel_plots}. The standard deviation of the Gaussian distribution was optimized separately for each forecasted time horizon. In each case, the standard deviation achieving the smallest ECE on the validation dataset was chosen. For the forecasted time horizons~$T_f$ and cell size~$e^2$ the standard deviations~$\sigma = [0.48 e, 0.48 e, 0.53 e, 0.55 e, 0.55 e]$ were found using model~d\_t for pedestrians and adopted for all discrete models. Due to the high computational costs required to optimize the standard deviations, the values were adopted for cyclists. An adjustment to the respective dataset could lead to further improvements. The procedure prevents overconfident distributions. However, the resulting reliability diagram shows S-shaped curves, especially for short forecasted time horizons. This effect is caused by discretization and will be discussed in more detail in~\cref{subsec_cell_size}. It should be noted here that reliable models can also be achieved using small network sizes, i.e., a small number of layers and neurons. But those networks are not able to make sharp forecasts at the same time because they are not able to learn the movement behavior precisely.
\begin{figure}
	\centering
		\begin{tabular}{cc}
			\begin{adjustbox}{clip,trim=0.1cm 0.0cm 0.05cm 0.0cm}\input{"images/results/calibration/reliability_plot_pedestrian_gesamt_no_gauss.pgf"}\end{adjustbox} &
			
			\begin{adjustbox}{clip,trim=0.8cm 0.0cm 0.05cm 0.0cm}\input{"images/results/calibration/reliability_plot_pedestrian_gesamt_optimal_gauss.pgf"}\end{adjustbox}
	\end{tabular}
	\caption{Reliability diagrams for model~d\_tp for pedestrians using plain cross entropy as loss function (left) and spatial label smoothing (right) evaluated on the validation dataset. The reliability is shown for the forecasted time horizons $\SI[round-mode=places,round-precision=2]{0.44}{\second}$ (black solid line), $\SI[round-mode=places,round-precision=2]{1.48}{\second}$ (dashed red line) and $\SI[round-mode=places,round-precision=2]{2.52}{\second}$ (dotted blue line). The ideal diagonal is plotted as dashed black line.}
	\label{fig_cal_rel_plots}
\end{figure}

\subsection{Comparison of the Various Methods}
\label{subsec_results}
In the following, the results of the final network configuration of each of the five models for pedestrians and cyclists are presented. All results are measured on the separate test dataset. Quantitative results following the evaluation metrics defined in~\cref{sec_eval_method} are provided for all models in~\cref{tab_ECE_ped,tab_S_ped,tab_ASWAEE_ped} for pedestrians and for cyclists in~\cref{tab_ECE_cyclists,tab_S_cyclists,tab_ASWAEE_cyclists}. Here, the results are distinguished for the different motion types. We discuss the impact of using poses and semantic maps on the forecasts and investigate the influence of the cell size on the forecasting results. Last but not least, we compare the discrete and continuous trajectory forecasting models.
\begin{table}
	\sisetup{round-mode=places,round-precision=1,scientific-notation=fixed, fixed-exponent=-2}
	\caption{Achieved ECE in $\si{\percent}$ of all models for pedestrians structured by motion types.}
	\begin{center}
		\begin{tabular}{p{0.6cm}*{5}{ S[table-format=1.2, table-omit-exponent]}}\toprule
			{model} & {all} & {wait} & {start} & {move} & {stop}\\\midrule
			c\_t & 0.049451110344247 & 0.124613904498992 & 0.044661770077223 & 0.044002816491892 & 0.041417986040101 \\
			d\_t & 0.071145601830397 & 0.059719220400471 & 0.095336800163726 & 0.067468592476097 & 0.102890935828377 \\[6pt]
			c\_tp & 0.03419160613344 & 0.082114809892202 & 0.033219999942261 & 0.031702043039779 & 0.041501626214362 \\
			d\_tp & 0.044777143792735 & 0.054206848146579 & 0.034752078820965 & 0.043357531576546 & 0.081579017844496 \\[6pt]
			d\_tpm & 0.03839612641079 & 0.068540987953754 & 0.036719817013977 & 0.038802364094418 & 0.063795291895465 \\\bottomrule
		\end{tabular}
		\label{tab_ECE_ped}
	\end{center}
\end{table}
\begin{table}
	\sisetup{round-mode=places,round-precision=2}
	\caption{Sharpness~$S(0.95)$ in {\upshape $\si{\meter\squared\per\second}$} of all models for pedestrians.}
	\begin{center}
		\begin{tabular}{p{0.6cm}*{5}{ S[table-format=1.2]}}\toprule
			{model} & {all} & {wait} & {start} & {move} & {stop}\\\midrule
			c\_t & 2.88916298301382 & 2.6549957323377 & 3.4298590597242 & 2.73660106613455 & 3.22750260124414 \\
			d\_t & 3.1078879731995 & 3.41400101736832 & 3.40800688349343 & 2.85609482962758 & 3.29593769946519 \\[6pt]
			c\_tp & 2.35003375321374 & 2.28466813586082 & 2.67077490147032 & 2.24066889349991 & 2.51636454170352 \\
			d\_tp & 3.03207209955285 & 2.98370435465402 & 3.32193004746299 & 2.93250525728722 & 3.16864389014796 \\[6pt]
			d\_tpm & 3.34101288953974 & 3.7243952317696 & 3.67670756539449 & 3.0607012604423 & 3.48938263301683 \\\bottomrule
			\end{tabular}
		\label{tab_S_ped}
	\end{center}
\end{table}
\begin{table}
	\sisetup{round-mode=places,round-precision=2}
	\caption{ASWAEE in {\upshape $\si{\meter\per\second}$} of all models for pedestrians.}
	\begin{center}
		\begin{tabular}{p{0.6cm}*{5}{ S[table-format=1.2]}}\toprule
			{model} & {all} & {wait} & {start} & {move} & {stop}\\\midrule
			c\_t & 0.568391086129987 & 0.507757098770849 & 0.653595733834182 & 0.5573663048427 & 0.606501403232862 \\
			d\_t & 0.603232608080936 & 0.543354598536583 & 0.666679840169642 & 0.597831427373795 & 0.642035111483152 \\[6pt]
			c\_tp & 0.513379840552458 & 0.495011270599892 & 0.55943715409892 & 0.498177972537536 & 0.547564363121977 \\
			d\_tp & 0.559968966512475 & 0.52780133051657 & 0.591422820742988 & 0.551209443513214 & 0.604954649753062 \\[6pt]
			d\_tpm & 0.571343102647215 & 0.551553895731079 & 0.616655434505315 & 0.55443342168419 & 0.614751244610632 \\\bottomrule
		\end{tabular}
		\label{tab_ASWAEE_ped}
	\end{center}
\end{table}
\begin{table}
	\sisetup{round-mode=places,round-precision=2,scientific-notation=fixed, fixed-exponent=-2}
	\caption{ECE in $\si{\percent}$ of all models for cyclists structured by motion types.}
	\begin{center}
		\begin{tabular}{p{0.6cm}*{7}{ S[table-format=1.2, table-omit-exponent]}}\toprule
			\multirow{2}{*}{model} & {\multirow{2}{*}{all}} & {\multirow{2}{*}{wait}} & {\multirow{2}{*}{start}} & {\multirow{2}{*}{move}} & {\multirow{2}{*}{stop}} & {turn} & {turn}\\
			& & & & & &{left} &{right}\\\midrule
			c\_t & 0.068280963843495 & 0.153404220680786 & 0.127128029635165 & 0.068658804281646 & 0.045312477101751 & 0.122632906648016 & 0.206761928231147 \\
			d\_t & 0.058985312346693 & 0.130637764064682 & 0.106304758297562 & 0.055765217329113 & 0.070395465443732 & 0.171950156558597 & 0.217767458425898 \\[6pt]
			c\_tp & 0.038411030755606 & 0.110698800765599 & 0.060497896955086 & 0.03224852800292 & 0.077999687440393 & 0.066827785090604 & 0.095060705674052 \\
			d\_tp & 0.075648539511195 & 0.1175460042118 & 0.094107113131143 & 0.075943519107756 & 0.081948967838753 & 0.191530173028818 & 0.214483865538248 \\[6pt]
			d\_tpm & 0.089910814724278 & 0.114610675417246 & 0.122721988422224 & 0.09093166296814 & 0.093014746989571 & 0.189899416484207 & 0.194914843184061 \\\bottomrule
		\end{tabular}
		\label{tab_ECE_cyclists}
	\end{center}
\end{table}
\begin{table}
	\sisetup{round-mode=places,round-precision=2}
	\caption{Sharpness~$S(0.95)$ in {\upshape $\si{\meter\squared\per\second}$} of all models for cyclists.}
	\begin{center}
		\begin{tabular}{p{0.6cm}*{7}{ S[table-format=1.2]}}\toprule
			\multirow{2}{*}{model} & {\multirow{2}{*}{all}} & {\multirow{2}{*}{wait}} & {\multirow{2}{*}{start}} & {\multirow{2}{*}{move}} & {\multirow{2}{*}{stop}} & {turn} & {turn}\\
			& & & & & &{left} &{right}\\\midrule
			c\_t & 4.715364570171658 & 1.7570619767005595 & 6.409134930920375 & 4.844837886715711 & 5.507069932709646 & 9.342498986636935 & 7.338185671282784\\
			d\_t & 4.60919755173473 & 2.48282917829234 & 6.60502085293957 & 4.65506444069618 & 5.56549220054106 & 8.19390170028741 & 7.5166771889745 \\[6pt]
			c\_tp & 5.378467342538193 & 2.5602421605144405 & 12.218712313380415 & 5.25618692045511 & 7.1401405165480085 & 14.919893751201688 & 14.350346551237712\\
			d\_tp & 3.47104257683254 & 2.01355529219987 & 5.35660586728305 & 3.4826007283011 & 4.13163291849724 & 5.86929375881356 & 5.87283563460124 \\[6pt]
			d\_tpm & 3.33283618613043 & 1.69957881608602 & 5.02505745720439 & 3.37379127029565 & 3.81085045161385 & 6.13833025626861 & 5.85077048346124 \\\bottomrule
			\end{tabular}
		\label{tab_S_cyclists}
	\end{center}
\end{table}
\begin{table}
	\sisetup{round-mode=places,round-precision=2}
	\caption{ASWAEE in {\upshape $\si{\meter\per\second}$} of all models for cyclists.}
	\begin{center}
		\begin{tabular}{p{0.6cm}*{7}{ S[table-format=1.2]}}\toprule
			\multirow{2}{*}{model} & {\multirow{2}{*}{all}} & {\multirow{2}{*}{wait}} & {\multirow{2}{*}{start}} & {\multirow{2}{*}{move}} & {\multirow{2}{*}{stop}} & {turn} & {turn}\\
			& & & & & &{left} &{right}\\\midrule
			c\_t & 0.680154470117705 & 0.37147463172119766 & 0.9339245169898645 & 0.6895377290297804 & 0.7884144841999351 & 1.0953898762940995 & 1.0847877779544541\\
			d\_t & 0.665051456534601 & 0.366673562932719 & 0.883156865558521 & 0.674346011258406 & 0.788198749203694 & 1.03783440180982 & 1.05749684552757 \\[6pt]
			c\_tp & 0.6392924611559083 & 0.4032345073463464 & 1.091070789275931 & 0.6336870936280591 & 0.7874615184977569 & 1.1751897504508813 & 1.146573140689958\\
			d\_tp & 0.632746692893105 & 0.371074413221297 & 0.806583405652432 & 0.64084833456761 & 0.756250972254148 & 0.993523478285246 & 1.00352645798216 \\[6pt]
			d\_tpm & 0.622393260388247 & 0.367537757614544 & 0.820262778760294 & 0.629502481231787 & 0.736080542898655 & 0.992696260781477 & 0.992247812729359 \\\bottomrule
		\end{tabular}
		\label{tab_ASWAEE_cyclists}
	\end{center}
\end{table}

\subsubsection{Impact of Poses}
\label{subsec_poses}
For pedestrians, with regard to positional accuracy, the use of poses results in a $\SI[round-mode=places,round-precision=1]{9.678}{\percent}$ reduction of the ASWAEE for the continuous forecasting approach~(motion type \textit{all} for models c\_t and c\_tp in~\cref{tab_ASWAEE_ped}). For the discrete method, an improvement of $\SI[round-mode=places,round-precision=1]{7.172}{\percent}$ is achieved~(motion type \textit{all} for models d\_t and d\_tp in~\cref{tab_ASWAEE_ped}). Meanwhile, improvements are also observed in the reliability~(\cref{tab_ECE_ped}) and sharpness~(\cref{tab_S_ped}) for both the discrete~(d\_t and d\_tp) and continuous~(c\_t and c\_tp) approaches. \cref{fig_comp_ped_poses} shows a comparison of discrete forecasts with and without the use of poses to exemplify their effects. The example in the top row shows the forecasts $\SI[round-mode=places,round-precision=2]{2.52}{\second}$ into the future of a pedestrian waiting at the curbside. The forecast based on the past trajectory (left) indicates continued standing as the most likely event and a low probability of starting in different directions. In comparison, the model~d\_tp~(right) can take the body orientation provided by the 3D poses into account. Accordingly, the forecast of a possible starting is made mainly towards the road. The second example~(lower row) presents a pedestrian who initially stands at the curbside and then crosses the road. In contrast to the trajectory based model, the model~d\_tp using poses detects the pedestrian's intention to start walking and forecasts a multimodal distribution covering a possible standing as well as crossing the street. This is an advantage of the discrete over the continuous methods used in this article, which cannot make multimodal forecasts. Overall, the forecasts for pedestrians are consistently improved by using poses for both the continuous and discrete approach. All motion types benefit from the additional information provided by poses. The improvement is indicated by better positional accuracy and reliability coupled with enhanced sharpness. 
\begin{figure}
	\centering
		\begin{tabular}{p{0.4\columnwidth}p{0.70\columnwidth}}
			\hspace{2.0cm}d\_t & \hspace{1.65cm}d\_tp \\
			\begin{adjustbox}{clip,trim=0.20cm 0.6cm 0.80cm 0.05cm}\input{"images/results/t_vs_tp/ped/t/1914_2_52.pgf"}\end{adjustbox} &   \begin{adjustbox}{clip,trim=0.5cm 0.6cm 0.1cm 0.05cm}\input{"images/results/t_vs_tp/ped/tp/1914_2_52.pgf"}\end{adjustbox} \\
			
			\begin{adjustbox}{clip,trim=0.20cm 0.0cm 0.80cm 0.05cm}\input{"images/results/t_vs_tp/ped/t/6143_2_52.pgf"}\end{adjustbox} &   \begin{adjustbox}{clip,trim=0.5cm 0.0cm 0.1cm 0.05cm}\input{"images/results/t_vs_tp/ped/tp/6143_2_52.pgf"}\end{adjustbox} \\
	\end{tabular}
	\caption{Comparison of two exemplary forecasts of the models d\_t~(left) and d\_tp~(right) for pedestrians and a forecasted time horizon of $\SI[round-mode=places,round-precision=2]{2.52}{\second}$. The forecasted distributions are illustrated using the logarithmic color scale on the right ranging from red (high probability) to blue (low probability). The past trajectory is represented by a white line/point and the actual position by a white cross. The semantic maps can be seen in the background. (Best viewed on screen).}
	\label{fig_comp_ped_poses}
\end{figure}

\begin{figure}
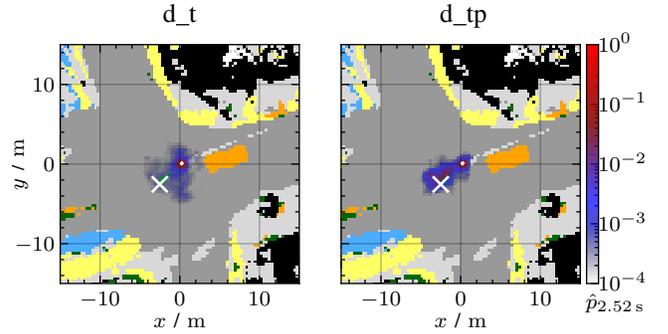

	\centering
		\begin{tabular}{p{0.4\columnwidth}p{0.70\columnwidth}}
			\hspace{2.0cm}d\_t & \hspace{1.65cm}d\_tp \\
			\begin{adjustbox}{clip,trim=0.20cm 0.0cm 0.80cm 0.05cm}\input{"images/results/t_vs_tp/bike/t/6704_2_52.pgf"}\end{adjustbox} &   \begin{adjustbox}{clip,trim=0.5cm 0.0cm 0.1cm 0.05cm}\input{"images/results/t_vs_tp/bike/tp/6704_2_52.pgf"}\end{adjustbox} \\
	\end{tabular}
	\caption{Comparison of exemplary forecasts of the models d\_t~(left) and d\_tp~(right) for cyclists and a forecasted time horizon of $\SI[round-mode=places,round-precision=2]{2.52}{\second}$.}
	\label{fig_comp_bike_poses}
\end{figure}

For cyclists, the ASWAEE is improved by $\SI[round-mode=places,round-precision=1]{6.007753533609439}{\percent}$ for the continuous method~(c\_t and c\_tp in~\cref{tab_ASWAEE_cyclists}) and $\SI[round-mode=places,round-precision=1]{4.857482830839632}{\percent}$ for the discrete approach~(d\_t and d\_tp in~\cref{tab_ASWAEE_cyclists}) by using poses. For the continuous model, the poses also lead to a considerable improvement in reliability~(\cref{tab_ECE_cyclists}) with slightly larger distributions with respect to the area~(\cref{tab_S_cyclists}). The area covered by a confidence level of $\SI{95}{\percent}$ is huge for turning cyclists since, in such situations, there is great variability and uncertainty. This is also evident from the high values obtained for the ASWAEE for these motion types~(\textit{turn left} and \textit{turn right} in~\cref{tab_ASWAEE_cyclists}). While the use of poses improves the positional accuracy~(\cref{tab_ASWAEE_cyclists}) for the discrete method~(model d\_tp vs d\_t), the reliability~(\cref{tab_ECE_cyclists}) is only enhanced for the motion types \textit{wait}, \textit{start}, and \textit{turn right}. For the remaining motion types, the reliability deteriorates as the model tends to forecast overconfident distributions. This is evident in low values for the sharpness~(d\_tp in~\cref{tab_S_cyclists}) and could be remedied by a separate optimization of the standard deviations for the spatial label smoothing. \Cref{fig_comp_bike_poses} presents an exemplary forecast for cyclists with (right) and without (left) poses. In this situation, a cyclist wants to turn left at an intersection. First, the cyclist stands in order to let oncoming traffic pass before initiating the turn. Similar to what we have seen with pedestrians, the model~d\_tp forecasts a multimodal distribution that covers continued standing as well as turning. The poses also provide the movement direction of the cyclist.

\subsubsection{Impact of Semantic Maps}
\label{sec_impact_semantic_maps}
\label{subsec_semantic_maps}
\begin{figure}
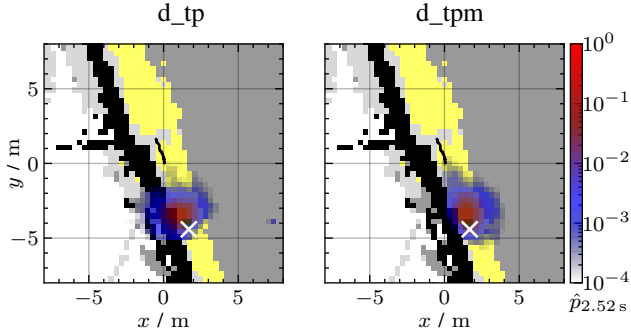

	\centering
		\begin{tabular}{p{0.4\columnwidth}p{0.70\columnwidth}}
			\hspace{2.0cm}d\_tp & \hspace{1.4cm}d\_tpm \\
			\begin{adjustbox}{clip,trim=0.20cm 0.0cm 0.80cm 0.05cm}\input{"images/results/tp_vs_tpc/ped/tp/10580_2_52.pgf"}\end{adjustbox} &   \begin{adjustbox}{clip,trim=0.5cm 0.0cm 0.1cm 0.05cm}\input{"images/results/tp_vs_tpc/ped/tpc/10580_2_52.pgf"}\end{adjustbox} \\
	\end{tabular}
	\caption{Comparison of exemplary forecasts of the models d\_tp~(left) and d\_tpm~(right) for pedestrians and a forecasted time horizon of $\SI[round-mode=places,round-precision=2]{2.52}{\second}$. For better visibility, the past trajectory is shown as black line here.}
	\label{fig_comp_ped_semantic_maps}
\end{figure}

\begin{figure}
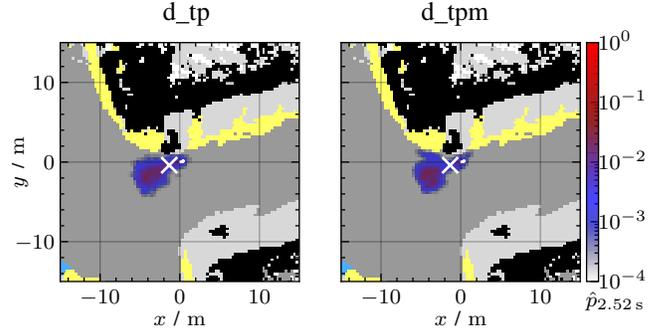

	\centering
		\begin{tabular}{p{0.4\columnwidth}p{0.70\columnwidth}}
			\hspace{2.0cm}d\_tp & \hspace{1.4cm}d\_tpm \\
			\begin{adjustbox}{clip,trim=0.20cm 0.0cm 0.80cm 0.05cm}\input{"images/results/tp_vs_tpc/bike/tp/7241_2_52.pgf"}\end{adjustbox} &   \begin{adjustbox}{clip,trim=0.5cm 0.0cm 0.1cm 0.05cm}\input{"images/results/tp_vs_tpc/bike/tpc/7241_2_52.pgf"}\end{adjustbox} \\
	\end{tabular}
	\caption{Comparison of exemplary forecasts of the models d\_tp~(left) and d\_tpm~(right) for cyclists and a forecasted time horizon of $\SI[round-mode=places,round-precision=2]{2.52}{\second}$.}
	\label{fig_comp_bike_semantic_maps}
\end{figure}

The semantic maps' impact on the forecasts is difficult to evaluate because no measure is known that allows for a quantification of the benefits. The meaning of the surrounding scene can be quite different from case to case. Therefore, identifying potential relations and the definition of reasonable measures are difficult. For pedestrians, the additional use of the semantic maps leads to better reliability~(d\_tpm in~\cref{tab_ECE_ped}) while the ASWAEE~(d\_tpm in~\cref{tab_ASWAEE_ped}) remains roughly the same compared to model~d\_tp. For cyclists, the ASWAEE also remains similar~(d\_tp vs d\_tpm in~\cref{tab_ASWAEE_cyclists}), while the reliability~(\cref{tab_ECE_cyclists}) deteriorates due to overconfident forecasts. However, these measures reflect the influence of semantic maps only to a limited extent. Therefore, in the following, we illustrate the impact of semantic maps qualitatively by some examples. In addition, we consider the occupancy score~$O_{t_f}$ as introduced in~\cref{sec_eval_method} since it allows the quantification of the influence of obstacles on the forecasts. \Cref{tab_O} reports the occupancy score for pedestrians and cyclists, the different discrete models, and each forecasted time horizon. For each model, the values for cyclists are smaller than those for pedestrians (e.g. row 3 vs row 6). This is plausible since cyclists are usually on the road and thus have a greater distance to static obstacles. Moreover, the generally small values in~\cref{tab_O} indicate a relatively small impact of the static obstacles on the forecasts in our dataset and the forecasted time horizons considered here. The score represents the average over the entire dataset. While for most pedestrians and cyclists, the future trajectory is sufficiently determined by the past trajectory and poses, obstacles are indeed important for individual examples. For the models~d\_t and d\_tp without use of the semantic maps, the score in~\cref{tab_O} rises as the forecasted time horizon grows, indicating an increase in the importance of the obstacles for the forecasts. The scores for all time horizons are reduced by considering the semantic maps for both pedestrians and cyclists~(d\_tpm in~\cref{tab_O}) which means that the forecasts of this model collide less with static obstacles. It should be noted that the optimal score is not necessarily zero, since the semantic maps contain errors that can be considered by the forecasting models. \Cref{fig_comp_ped_semantic_maps} illustrates the forecasts of model~d\_tp~(left) and d\_tpm~(right) of a pedestrian walking on the sidewalk alongside a house wall. While the forecasted distribution of model~d\_tp is approximately symmetric with respect to the movement direction, model~d\_tpm considers the house wall resulting in a skewed distribution. For cyclists, model~d\_tpm in particular takes into account the course of the road and the curbside. An example of this can be seen in~\cref{fig_comp_bike_semantic_maps}. It shows a cyclist who is just entering an intersection. In addition to driving straight and turning left, model~d\_tpm also forecasts a low probability for turning right following the curb.
\begin{table}
    \sisetup{round-mode=places,round-precision=2,scientific-notation=fixed, fixed-exponent=-2}
	\caption{Occupancy Score~$O_{t_f}$ in \% of all discrete models for different forecasted time horizons and VRU types.}
	\begin{center}
		\begin{tabular}{p{1.2cm} p{0.6cm}*{5}{ S[table-format=1.2, table-omit-exponent]}}\toprule
			& model & {0.44\,s} & {0.96\,s} & {1.48\,s} & {2.00\,s} & {2.52\,s} \\\midrule
			\multirow{ 3}{*}{pedestrians} & {d\_t} & 0.003906333 & 0.0058334074 & 0.008984631 & 0.013530846 & 0.01992912 \\
			& {d\_tp} & 0.003746882 & 0.0052810824 & 0.00774657 & 0.011179768 & 0.015829882 \\
			& {d\_tpm} & 0.0025997653 & 0.0036633357 & 0.0047773756 & 0.0057684015 & 0.0068251407 \\[6pt]
			\multirow{ 3}{*}{cyclists} & {d\_t} & 0.00126393 & 0.0016439919 & 0.0027795932 & 0.0040555173 & 0.006403441 \\
			& {d\_tp} & 0.0013422493 & 0.0015665151 & 0.0026805666 & 0.0040352102 & 0.006307096 \\
			& {d\_tpm} & 0.00094635715 & 0.0012465471 & 0.0016948616 & 0.0016605583 & 0.001652842 \\\bottomrule
		\end{tabular}
		\label{tab_O}
	\end{center}
\end{table}

\subsubsection{Impact of Cell Size}
\label{subsec_cell_size}
To investigate the effects of the grid cell size, the model~d\_tp is trained for pedestrians with a cell size of $e^2=0.175\times\SI[round-mode=places,round-precision=3]{0.175}{\metre\squared}$ and compared with the previously reported results using a cell size of $0.35\times\SI[round-mode=places,round-precision=2]{0.35}{\metre\squared}$. Increasing the resolution offers two potential advantages: First, the model can provide a finer resolution of the forecasted distribution, and second, the actual position used to train the model is also resolved more precisely. Both can have a beneficial effect on the forecasting results. However, this comes at the cost of increased demand for computational and memory resources. The training time for a single batch of size~$40$ increases from $\SI[round-mode=places,round-precision=1]{7.1}{\milli\second}$ to $\SI[round-mode=places,round-precision=1]{18.6}{\milli\second}$ using an Nvidia~RTX~2080~Ti and the required graphics memory from $\SI[round-mode=places,round-precision=1]{1.755}{\giga\byte}$ to $\SI[round-mode=places,round-precision=1]{4.977}{\giga\byte}$. The increase is even greater for cyclists due to the larger grid: Here, the training time for a single batch increases from $\SI[round-mode=places,round-precision=1]{21.9}{\milli\second}$ to $\SI[round-mode=places,round-precision=1]{83.3}{\milli\second}$ and the needed memory from $\SI[round-mode=places,round-precision=1]{3.100}{\giga\byte}$ to $\SI[round-mode=places,round-precision=1]{10.347}{\giga\byte}$. As a result, a detailed examination of the effects of the grid cell size is not possible for cyclists and we limit ourselves to pedestrians. However, it should be noted here that the cell size has little effect on the inference time: For pedestrians, the inference time remains at $\SI[round-mode=places,round-precision=1]{1.1}{\milli\second}$ for both resolutions, while for cyclists it increases from $\SI[round-mode=places,round-precision=1]{1.1}{\milli\second}$ to $\SI[round-mode=places,round-precision=1]{1.6}{\milli\second}$. This shows that the high resource demand for a fine resolution is required mainly during training and not during the use of the models.

\Cref{tab_ped_hd} shows the results for pedestrians obtained using the smaller cell size. Compared to the larger cell size (given in parentheses in~\cref{tab_ped_hd}), there is a considerable improvement in all scores. This model even outperforms the continuous model~c\_tp in terms of reliability~(c\_tp in \cref{tab_ECE_ped} vs \cref{tab_ped_hd}) while achieving slightly worse results for the ASWAEE~(c\_tp in \cref{tab_ASWAEE_ped} vs \cref{tab_ped_hd}). A look at the corresponding reliability diagram on the right in~\cref{fig_hd_rel_plots} reveals the main reason for the improvement in reliability. Compared to the reliability diagram of model~d\_tp with the larger cell size (on the left in~\cref{fig_hd_rel_plots}), the S-shape of the curves is reduced. This particularly affects short forecasted time horizons, since for them, the forecasted distributions are spread over relatively small areas.

\begin{table}
	\def\x#1{~(\SI[round-mode=places,round-precision=2,scientific-notation=false]{#1}{})}
	\caption{The achieved results of the model~{\upshape d\_tp} for pedestrians using a cell size of {\upshape $0.175\times\SI[round-mode=places,round-precision=3]{0.175}{\metre\squared}$} for each motion type. For comparison, the results of the same model for cell size {\upshape $0.35\times\SI[round-mode=places,round-precision=2]{0.35}{\metre\squared}$} are provided in parentheses.}
	\begin{center}
		\begin{tabular}{p{1.4cm} S[round-mode=places,round-precision=2,scientific-notation=fixed, fixed-exponent=-2, table-format=1.2, table-omit-exponent] S[table-format=1.2, round-mode=places,round-precision=2] S[round-mode=places,round-precision=2, table-format=1.2]}\toprule
			{motion type} & {~ECE in $\si{\percent}$} & {$S(0.95)$ in $\si{\meter\squared\per\second}$} & {ASWAEE in $\si{\meter\per\second}$} \\\midrule
			all & 0.0298979610316 \x{4.4777143792735} & 2.80969592189404 \x{3.03207209955285} & 0.527254330595919 \x{0.559968966512475} \\
			wait & 0.029017860331862 \x{5.4206848146579} & 3.04867668215767 \x{2.98370435465402} & 0.487956394532791 \x{0.52780133051657} \\
			start & 0.049357790033629 \x{3.4752078820965} & 3.03851439486237 \x{3.32193004746299} & 0.57730638580387 \x{0.591422820742988} \\
			move & 0.024417692751583 \x{4.3357531576546} & 2.62856394900354 \x{2.93250525728722} & 0.514899706791391 \x{0.551209443513214} \\
			stop & 0.077132154100308 \x{8.1579017844496} & 2.90578284994658 \x{3.16864389014796} & 0.576405225678724 \x{0.604954649753062} \\\bottomrule
		\end{tabular}
		\label{tab_ped_hd}
	\end{center}
\end{table}

\begin{figure}
	\centering
		\begin{tabular}{cc}
			\begin{adjustbox}{clip,trim=0.1cm 0.0cm 0.05cm 0.0cm}\input{"images/results/high_resolution/tp/reliability_plot_pedestrian_gesamt.pgf"}\end{adjustbox} &
			
			\begin{adjustbox}{clip,trim=0.8cm 0.0cm 0.05cm 0.0cm}\input{"images/results/high_resolution/tp_hd/reliability_plot_pedestrian_gesamt.pgf"}\end{adjustbox}
	\end{tabular}
	\caption{Reliability diagrams for model~d\_tp and pedestrians using spatial label smoothing and a cell size of $0.35\times\SI[round-mode=places,round-precision=2]{0.35}{\metre\squared}$~(left) and $0.175\times\SI[round-mode=places,round-precision=3]{0.175}{\metre\squared}$~(right) evaluated on the test dataset. The reliability is shown for the time horizons $\SI[round-mode=places,round-precision=2]{0.44}{\second}$ (black solid line), $\SI[round-mode=places,round-precision=2]{1.48}{\second}$ (dashed red line) and $\SI[round-mode=places,round-precision=2]{2.52}{\second}$ (dotted blue line). The ideal diagonal is plotted as dashed black line.}
	\label{fig_hd_rel_plots}
\end{figure}

\subsubsection{Comparison of Discrete and Continuous Forecasting Models}
\label{subsec_comparison}
Overall, the continuous models~c\_t and c\_tp achieve better results for pedestrians in terms of reliability~(\cref{tab_ECE_ped}), sharpness~(\cref{tab_S_ped}), and ASWAEE~(\cref{tab_ASWAEE_ped}) compared to the discrete models~d\_t and d\_tp with cell size $0.35\times\SI[round-mode=places,round-precision=2]{0.35}{\metre\squared}$. In contrast, for cyclists, using solely the past trajectory as input, the discrete model achieves better scores~(c\_t vs d\_t in~\cref{tab_ECE_cyclists,tab_S_cyclists,tab_ASWAEE_cyclists}). The use of poses and semantic maps~(d\_tp and d\_tpm), on the other hand, leads to overconfident forecasts so that the reliability of the continuous model~c\_tp is in front. Both approaches have specific advantages and disadvantages: The used continuous method forecasts a single Gaussian distribution, so the type of the distribution is fixed and multimodal distributions cannot be represented. This disadvantage can be observed, for example, in the reliability of waiting pedestrians~(c\_tp vs d\_tp for \textit{wait} in~\cref{tab_ECE_ped}). Here, the continuous method~c\_tp cannot express the unlikely but possible chance of a starting motion. The discrete method~d\_tp has no such limitations. As the examples have demonstrated, it can express unsymmetric and multimodal distributions. However, this comes at the cost of discretizing the space leading to less accurate forecasts. The positions of the VRUs can only be forecasted within the chosen grid, and the required computational resources for the training limit the cell size. The use of the cross entropy as loss function leads to overconfident distributions. It can be compensated by spatial label smoothing, but the determination of the necessary parameters is a tedious process. On the other hand, the discrete method allows the consideration of the surroundings in the form of semantic maps. Their impact depends on the particular situation and the forecasted time horizon. There is no continuous approach in the literature allowing the consideration of semantic maps. It is difficult for continuous methods to learn and predict arbitrary distributions and to adjust them to the semantic maps.

	\section{\large Conclusions and Future Work}
\label{sec_conclusion}
In this article, we presented an approach for probabilistic trajectory forecasting of pedestrians and cyclists considering past movements represented by 3D poses and the surroundings in the form of semantic maps. The forecasts are generated in discrete grids allowing the model to learn arbitrary distributions. The impact of poses and semantic maps on the forecasts was examined, and the forecasted distributions were evaluated by their reliability, sharpness, and positional accuracy. We compared our approach with a method for forecasting continuous Gaussian distributions discussing their respective advantages and disadvantages.
Using the plain cross entropy as loss function leads to overconfident distributions. This can be prevented by applying spatial label smoothing. The resolution of the grids has a non-negligible influence on the quality of the results. Here, a trade-off must be made, taking into account the required computational and memory resources. Overall, the use of 3D poses improves the forecasts, especially through better detection of motion type changes and body orientation. To adequately forecast the behavior of VRUs, multimodal and skewed distributions are advantageous which are possible with the proposed method. The semantic maps allow a precise adaptation of the forecasts to the individual situation. While this leads to major improvements of the forecasts in individual cases, the semantic maps have an overall subordinate impact on the entire dataset and the forecasted time horizon of $\SI[round-mode=places,round-precision=2]{2.52}{\second}$. However, their impact increases for longer time horizons. Therefore, the surroundings must be taken into account in long-term forecasting.

Our future work will focus on using our approach to model specific motion types to improve the forecasts. A preceding motion type detection should be used to weigh the individual models. Furthermore, we want to improve the quality of the semantic maps contributing to better trajectory forecasts. The impact of considering semantic maps in motion type detection of VRUs will also be investigated.
	
	\section*{Acknowledgment}
This work was supported by ``Zentrum Digitalisierung.Bayern''. 
In addition, the work is backed by the project DeCoInt$^2$, supported by the German Research Foundation (DFG) within the priority program SPP 1835: ``Kooperativ interagierende Automobile'', grant numbers DO 1186/1-2 and SI 674/11-2.
	
	
	
	\bibliographystyle{IEEEtran}
	\bibliography{mybibfile}

\begin{thebibliography}{10}
\providecommand{\url}[1]{#1}
\csname url@samestyle\endcsname
\providecommand{\newblock}{\relax}
\providecommand{\bibinfo}[2]{#2}
\providecommand{\BIBentrySTDinterwordspacing}{\spaceskip=0pt\relax}
\providecommand{\BIBentryALTinterwordstretchfactor}{4}
\providecommand{\BIBentryALTinterwordspacing}{\spaceskip=\fontdimen2\font plus
\BIBentryALTinterwordstretchfactor\fontdimen3\font minus
  \fontdimen4\font\relax}
\providecommand{\BIBforeignlanguage}[2]{{%
\expandafter\ifx\csname l@#1\endcsname\relax
\typeout{** WARNING: IEEEtran.bst: No hyphenation pattern has been}%
\typeout{** loaded for the language `#1'. Using the pattern for}%
\typeout{** the default language instead.}%
\else
\language=\csname l@#1\endcsname
\fi
#2}}
\providecommand{\BIBdecl}{\relax}
\BIBdecl

\bibitem{Keller.2014}
C.~Keller and D.~Gavrila, ``Will the pedestrian cross? a study on pedestrian
  path prediction,'' \emph{IEEE Transactions on Intelligent Transportation
  Systems}, vol.~15, no.~2, pp. 494--506, 2014.

\bibitem{Goldhammer.2019}
M.~{Goldhammer}, S.~{K\"ohler}, S.~{Zernetsch}, K.~{Doll}, B.~{Sick}, and
  K.~{Dietmayer}, ``Intentions of vulnerable road users-—detection and
  forecasting by means of machine learning,'' \emph{IEEE Transactions on
  Intelligent Transportation Systems}, vol.~21, no.~7, pp. 3035--3045, 2020.

\bibitem{Gupta.2018}
A.~Gupta, J.~Johnson, L.~Fei-Fei, S.~Savarese, and A.~Alahi, ``Social {GAN}:
  Socially acceptable trajectories with generative adversarial networks,'' in
  \emph{{IEEE}/{CVF} Conference on Computer Vision and Pattern Recognition},
  2018, pp. 2255--2264.

\bibitem{Marchetti.2020}
F.~Marchetti, F.~Becattini, L.~Seidenari, and A.~D. Bimbo, ``{MANTRA}: Memory
  augmented networks for multiple trajectory prediction,'' in
  \emph{{IEEE}/{CVF} Conference on Computer Vision and Pattern Recognition
  ({CVPR})}, 2020, pp. 7141--7150.

\bibitem{Alahi.2016}
A.~{Alahi}, K.~{Goel}, V.~{Ramanathan}, A.~{Robicquet}, L.~{Fei-Fei}, and
  S.~{Savarese}, ``Social lstm: Human trajectory prediction in crowded
  spaces,'' in \emph{IEEE Conference on Computer Vision and Pattern Recognition
  (CVPR)}, 2016, pp. 961--971.

\bibitem{pool.2019}
E.~A.~I. Pool, J.~F.~P. Kooij, and D.~M. Gavrila, ``Context-based cyclist path
  prediction using {Recurrent} {Neural} {Networks},'' in \emph{{IEEE}
  {Intelligent} {Vehicles} {Symposium} ({IV})}, 2019, pp. 824--830.

\bibitem{Eilbrecht.2017}
J.~{Eilbrecht}, M.~{Bieshaar}, S.~{Zernetsch}, K.~{Doll}, B.~{Sick}, and
  O.~{Stursberg}, ``Model-predictive planning for autonomous vehicles
  anticipating intentions of vulnerable road users by artificial neural
  networks,'' in \emph{IEEE Symposium Series on Computational Intelligence
  (SSCI)}, 2017, pp. 1--8.

\bibitem{Koschi.2018}
M.~{Koschi}, C.~{Pek}, M.~{Beikirch}, and M.~{Althoff}, ``Set-based prediction
  of pedestrians in urban environments considering formalized traffic rules,''
  in \emph{IEEE Intelligent Transportation Systems Conference (ITSC)}, 2018,
  pp. 2704--2711.

\bibitem{Zernetsch.2019}
S.~{Zernetsch}, H.~{Reichert}, V.~{Kress}, K.~{Doll}, and B.~{Sick},
  ``Trajectory forecasts with uncertainties of vulnerable road users by means
  of neural networks,'' in \emph{IEEE Intelligent Vehicles Symposium (IV)},
  2019, pp. 810--815.

\bibitem{Bieshaar.2020}
M.~Bieshaar, J.~Schreiber, S.~Vogt, A.~Gensler, and B.~Sick, ``Quantile
  surfaces -- generalizing quantile regression to multivariate targets,''
  arXiv: 2010.05898, \url{https://arxiv.org/abs/2010.05898}, 2020.

\bibitem{Wu.2018}
J.~Wu, J.~Ruenz, and M.~Althoff, ``Probabilistic {Map}-based {Pedestrian}
  {Motion} {Prediction} {Taking} {Traffic} {Participants} into
  {Consideration},'' in \emph{{IEEE} {Intelligent} {Vehicles} {Symposium}
  ({IV})}, 2018, pp. 1285--1292.

\bibitem{Jain.2019}
A.~Jain, S.~Casas, R.~Liao, Y.~Xiong, S.~Feng, S.~Segal, and R.~Urtasun,
  ``Discrete {Residual} {Flow} for {Probabilistic} {Pedestrian} {Behavior}
  {Prediction},'' in \emph{Conference on {Robot} {Learning} (CoRL)}, ser.
  Proceedings of Machine Learning Research, vol. 100, 2019, pp. 407--419.

\bibitem{Quintero.2019}
R.~Quintero, I.~Parra, D.~Fern\'andez-Llorca, and M.~A. {Sotelo}, ``Pedestrian
  path, pose, and intention prediction through gaussian process dynamical
  models and pedestrian activity recognition,'' \emph{IEEE Transactions on
  Intelligent Transportation Systems}, vol.~20, no.~5, pp. 1803--1814, 2019.

\bibitem{Kress.2019c}
V.~Kress, S.~Zernetsch, K.~Doll, and B.~Sick, ``Pose based trajectory forecast
  of vulnerable road users,'' in \emph{IEEE Symposium Series on Computational
  Intelligence (SSCI)}, 2019, pp. 1200--1207.

\bibitem{Xue.2018}
H.~Xue, D.~Q. Huynh, and M.~Reynolds, ``{SS}-{LSTM}: {A} {Hierarchical} {LSTM}
  {Model} for {Pedestrian} {Trajectory} {Prediction},'' in \emph{{IEEE}
  {Winter} {Conference} on {Applications} of {Computer} {Vision} ({WACV})},
  2018, pp. 1186--1194.

\bibitem{Cheng.2020}
H.~Cheng, W.~Liao, M.~Y. Yang, M.~Sester, and B.~Rosenhahn, ``{MCENET}:
  {Multi}-{Context} {Encoder} {Network} for {Homogeneous} {Agent} {Trajectory}
  {Prediction} in {Mixed} {Traffic},'' in \emph{{IEEE} {Intelligent}
  {Transportation} {Systems} {Conference} ({ITSC})}, 2020, pp. 1--8.

\bibitem{Ridel.2020}
D.~Ridel, N.~Deo, D.~Wolf, and M.~Trivedi, ``Scene {Compliant} {Trajectory}
  {Forecast} {With} {Agent}-{Centric} {Spatio}-{Temporal} {Grids},'' \emph{IEEE
  Robotics and Automation Letters}, vol.~5, no.~2, pp. 2816--2823, 2020.

\bibitem{Kress_Dataset.2021}
\BIBentryALTinterwordspacing
V.~Kress, S.~Zernetsch, M.~Bieshaar, G.~Reitberger, E.~Fuchs, K.~Doll, and
  B.~Sick, ``{Pedestrians and Cyclists in Road Traffic: Trajectories, 3D Poses
  and Semantic Maps},'' 2021. [Online]. Available:
  \url{https://doi.org/10.5281/zenodo.4898838}
\BIBentrySTDinterwordspacing

\bibitem{Cao.2017}
Z.~Cao, T.~Simon, S.-E. Wei, and Y.~Sheikh, ``Realtime multi-person 2d pose
  estimation using part affinity fields,'' in \emph{IEEE Conference on Computer
  Vision and Pattern Recognition (CVPR)}, 2017, pp. 1302--1310.

\bibitem{Tome.2017}
D.~Tome, C.~Russell, and L.~Agapito, ``Lifting from the deep: Convolutional 3d
  pose estimation from a single image,'' in \emph{IEEE Conference on Computer
  Vision and Pattern Recognition (CVPR)}, 2017, pp. 5689--5698.

\bibitem{Kress.2018}
V.~Kress, J.~Jung, S.~Zernetsch, K.~Doll, and B.~Sick, ``Human pose estimation
  in real traffic scenes,'' in \emph{IEEE Symposium Series on Computational
  Intelligence (SSCI)}, 2018, pp. 518--523.

\bibitem{Wu.2019}
Z.~Wu, C.~Shen, and A.~van~den Hengel, ``Wider or {Deeper}: {Revisiting} the
  {ResNet} {Model} for {Visual} {Recognition},'' \emph{Pattern Recognition},
  vol.~90, pp. 119--133, 2019.

\bibitem{Ronneberger.2015}
O.~Ronneberger, P.~Fischer, and T.~Brox, ``U-net: Convolutional networks for
  biomedical image segmentation,'' in \emph{Medical Image Computing and
  Computer-Assisted Intervention – {MICCAI}}, ser. Lecture Notes in Computer
  Science, 2015, pp. 234--241.

\bibitem{Shi.2015}
X.~Shi, Z.~Chen, H.~Wang, D.~Yeung, W.~Wong, and W.~Woo, ``Convolutional {LSTM}
  network: {A} machine learning approach for precipitation nowcasting,'' in
  \emph{International Conference on Neural Information Processing Systems
  (NIPS)}, 2015, pp. 802--810.

\bibitem{Naeini.2015}
M.~P. Naeini, G.~F. Cooper, and M.~Hauskrecht, ``Obtaining {Well} {Calibrated}
  {Probabilities} {Using} {Bayesian} {Binning},'' in \emph{AAAI Conference on
  Artificial Intelligence}, 2015, pp. 2901--2907.

\bibitem{Murphy.1973}
A.~H. Murphy, ``A {New} {Vector} {Partition} of the {Probability} {Score},''
  \emph{Journal of Applied Meteorology (1962-1982)}, vol.~12, no.~4, pp.
  595--600, 1973.

\bibitem{Kingma.2014}
D.~P. Kingma and J.~Ba, ``Adam: A method for stochastic optimization,'' in
  \emph{International Conference on Learning Representations (ICLR)}, 2015.

\bibitem{Guo.2017}
C.~Guo, G.~Pleiss, Y.~Sun, and K.~Q. Weinberger, ``On {Calibration} of {Modern}
  {Neural} {Networks},'' in \emph{International {Conference} on {Machine}
  {Learning}}.\hskip 1em plus 0.5em minus 0.4em\relax PMLR, 2017, pp.
  1321--1330.

\end{thebibliography}
	
	
	\begin{IEEEbiography}[{\includegraphics[width=1in,height=1.25in,clip,keepaspectratio]{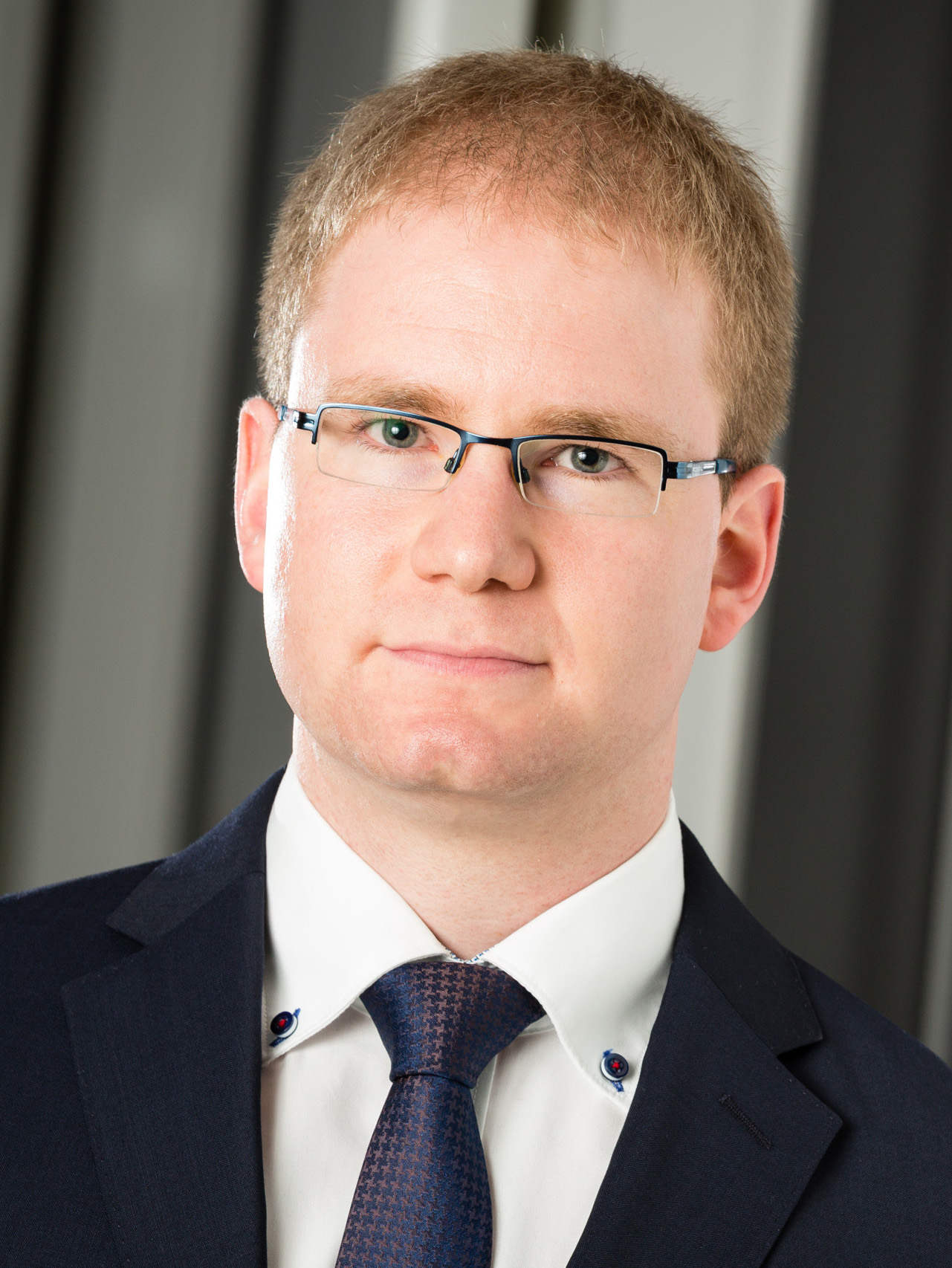}}]{Viktor Kress}
	received the B.Eng. degree in Mechatronics and the M.Eng. degree in Electrical Engineering and Information Technology from the University of Applied Sciences Aschaffenburg, Germany, in 2015 and 2016, respectively. Currently, he is working on his PhD thesis in cooperation with the Faculty of Electrical Engineering and Computer Science of the University of Kassel, Germany. His research interests include sensor data fusion, pattern recognition, machine learning, behavior recognition and trajectory forecasting of traffic participants.
\end{IEEEbiography}
\vspace{-10mm}

\begin{IEEEbiography}[{\includegraphics[width=1in,height=1.25in,clip,keepaspectratio]{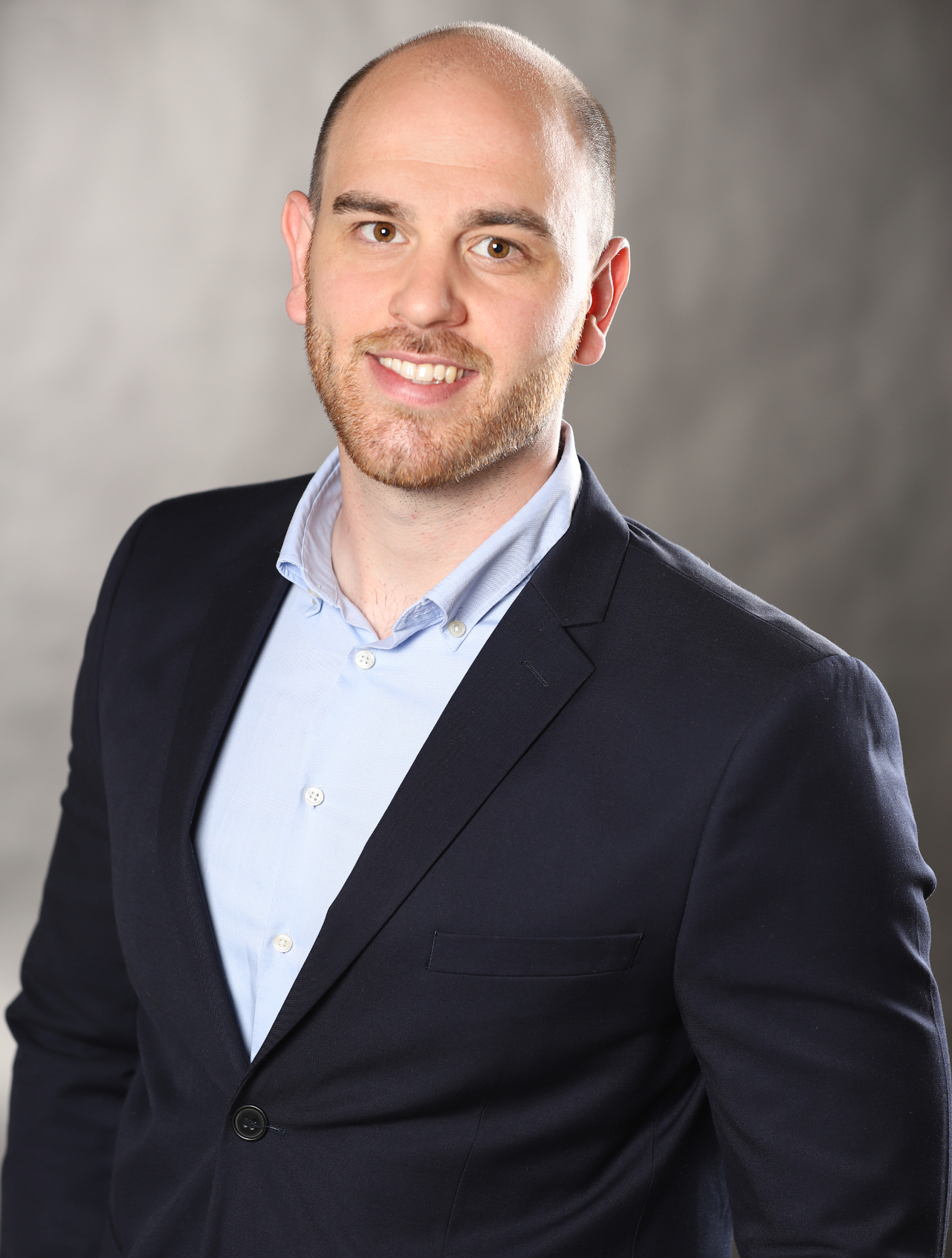}}]{Fabian Jeske}
	received a B.Eng. degree Industrial Engineering from the University of Applied Sciences Aschaffenburg, Germany in 2019. At the moment he is working on his M.Sc. in Industrial Engineering at the University of Applied Sciences Aschaffenburg. His research interests include trajectory forecasting of traffic participants, sensor data fusion and pattern recognition.
\end{IEEEbiography}
\vspace{-10mm}

\begin{IEEEbiography}[{\includegraphics[width=1in,height=1.25in,clip,keepaspectratio]{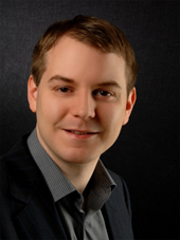}}]{Stefan Zernetsch}
	received the B.Eng. and the M.Eng. degree in Electrical Engineering and Information Technology from the University of Applied Sciences Aschaffenburg, Germany, in 2012 and 2014, respectively. Currently, he is working on his PhD thesis in cooperation with the Faculty of Electrical Engineering and Computer Science of the University of Kassel, Germany. His research interests include cooperative sensor networks, data fusion, multiple view geometry, pattern recognition and behavior recognition of traffic participants.
\end{IEEEbiography}
\vspace{-10mm}

\begin{IEEEbiography}[{\includegraphics[width=1in,height=1.25in,clip,keepaspectratio]{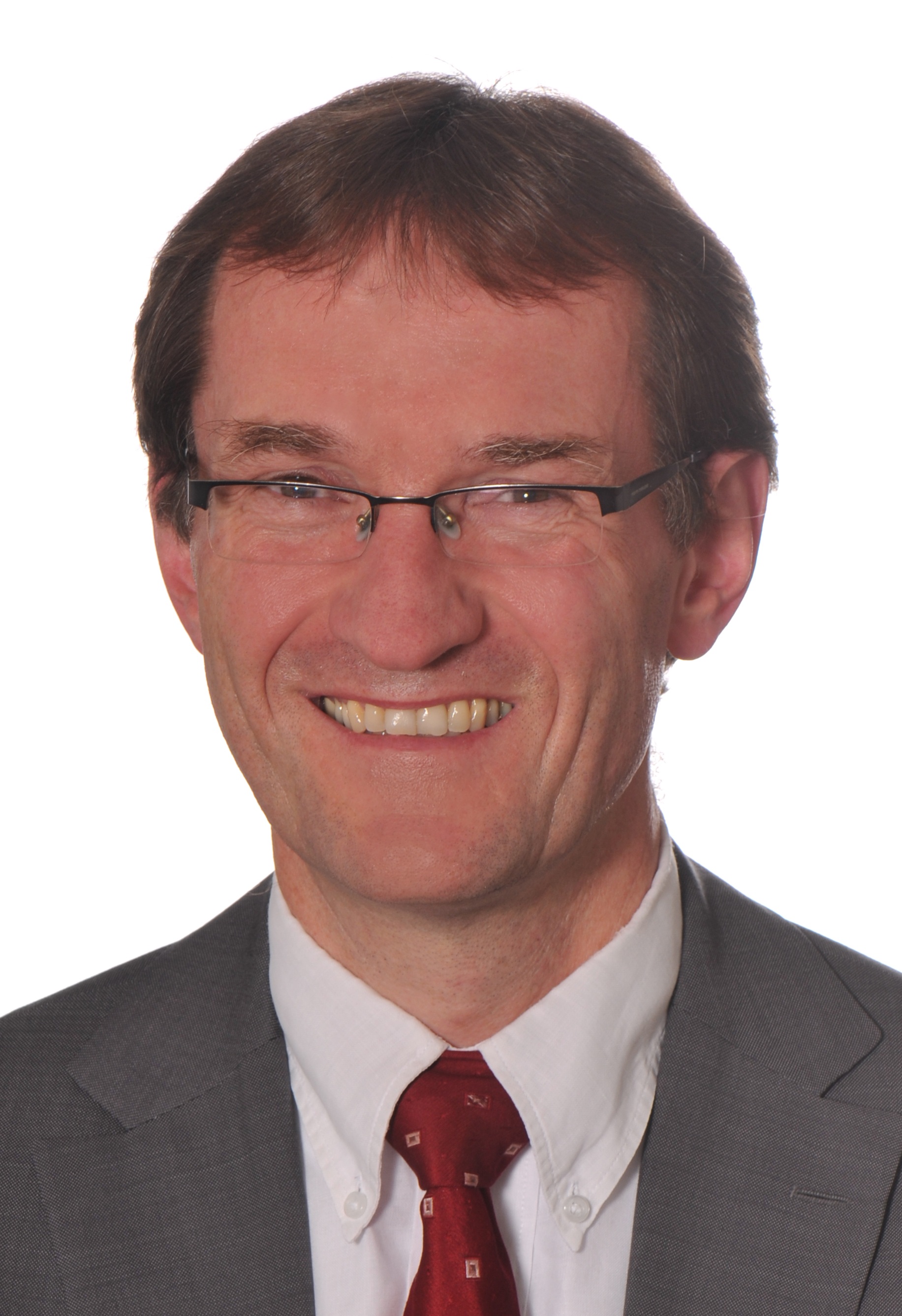}}]{Konrad Doll}
	received the Diploma (Dipl.-Ing.) degree and the Dr.-Ing. degree in Electrical Engineering and Information Technology from the Technical University of Munich, Germany, in 1989 and 1994, respectively. In 1994 he joined the Semiconductor Products Sector of Motorola, Inc. (now Freescale Semiconductor, Inc.). In 1997 he was appointed to professor at the University of Applied Sciences Aschaffenburg in the field of computer science and digital systems design. His research interests include intelligent systems, their real-time implementations, and their applications in advanced driver assistance systems and automated driving. He received several thesis and best paper awards. Konrad Doll is member of the IEEE.
\end{IEEEbiography}
\vspace{-10mm}
\begin{IEEEbiography}[{\includegraphics[width=1in,height=1.25in,clip,keepaspectratio]{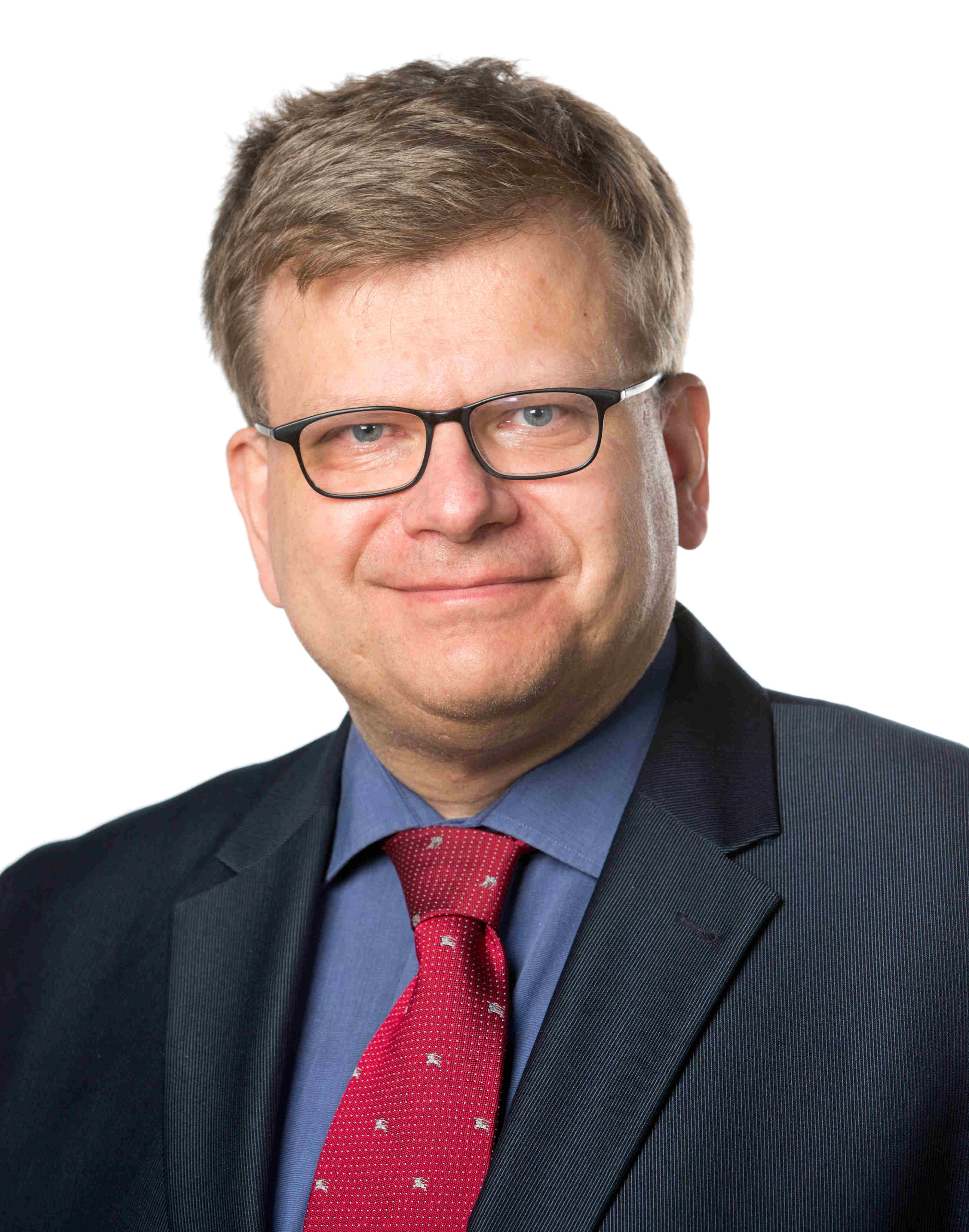}}]{Bernhard Sick}
	received the diploma, the Ph.D. degree, and the "Habilitation" degree, all in computer science, from the University of Passau, Germany, in 1992, 1999, and 2004, respectively. Currently, he is full Professor for Intelligent Embedded Systems at the Faculty for Electrical Engineering and Computer Science of the University of Kassel, Germany. There, he is conducting research in the areas autonomic and organic computing and technical data analytics with applications in, e.g., energy systems, automotive engineering, physics and materials science. He authored more than 200 peer-reviewed publications in these areas. Dr. Sick is associate editor of the IEEE TRANSACTIONS ON CYBERNETICS. He holds one patent and received several thesis, best paper, teaching, and inventor awards. He is a member of IEEE (Systems, Man, and Cybernetics Society, Computer Society, and Computational Intelligence Society) and GI (Gesellschaft fuer Informatik).
\end{IEEEbiography}

\end{document}